\documentclass{article}
\usepackage{amssymb}

\usepackage[final]{corl_2025} %
\usepackage{times}

\usepackage[numbers]{natbib}
\usepackage{subcaption}
\usepackage{graphicx}
\usepackage{lipsum}  
\usepackage[table]{xcolor}
\usepackage{makecell}
\usepackage{amsmath}
\usepackage{multirow}
\usepackage{booktabs}
\usepackage{array}
\usepackage{etoolbox}

\usepackage{wrapfig}
\usepackage{amsfonts}
\usepackage[shortlabels]{enumitem} %
\usepackage{twemojis}
\usepackage{bbm}

\newcommand{\cameraready}[1]{\textcolor{black}{#1}}

\usepackage{mathtools, algorithm, algpseudocode}
\usepackage{cleveref}

\definecolor{hiroshige}{HTML}{ffd06f}
\definecolor{algblue}{RGB}{0, 112, 192}
\definecolor{darkBlue}{RGB}{10,50,220}

\definecolor{darkBlue}{RGB}{10,50,220}
\definecolor{customRed}{RGB}{190,110,113}
\definecolor{customGreen}{RGB}{70,170,80}

\hypersetup{
	colorlinks=true,
	linkcolor=customGreen,
	filecolor=magenta,      
	urlcolor=cyan,
	citecolor=customRed,
}

\newcommand{\policy}{\pi_\theta (a_t \mid o_t, z)}
\newcommand{\rewardgeneral}{R_\psi(o_{1:t}, z)}
\newcommand{\reward}{\rewardgeneral}

\newcommand{\lossprogress}{\mathcal{L}_\text{progress}(o_{1:T}, z, o_{1:T}^\text{other})}
\newcommand{\success}{r_\text{success}}
\newcommand{\demos}{\mathcal{D}_\text{demos}}
\newcommand{\openx}{\mathcal{D}_\text{open-x}}

\newcommand{\method}{ReWiND}
\newcommand{\methodlong}{\method\ (\textbf{Re}wards \textbf{Wi}thout \textbf{N}ew \textbf{D}emonstrations)}
\newcommand{\outputhead}{cross-modal sequential aggregator}
\newcommand{\lossrewind}{\mathcal{L}_\text{rewind}(o_{1:T}, z)}

\newcommand{\onlinerewards}{\hat{r}^\text{on}}
\newcommand{\offlinerewards}{\hat{r}^\text{off}}
\newcommand{\newtext}[1]{\textcolor{black}{#1}}
\title{
\includegraphics[height=7mm]{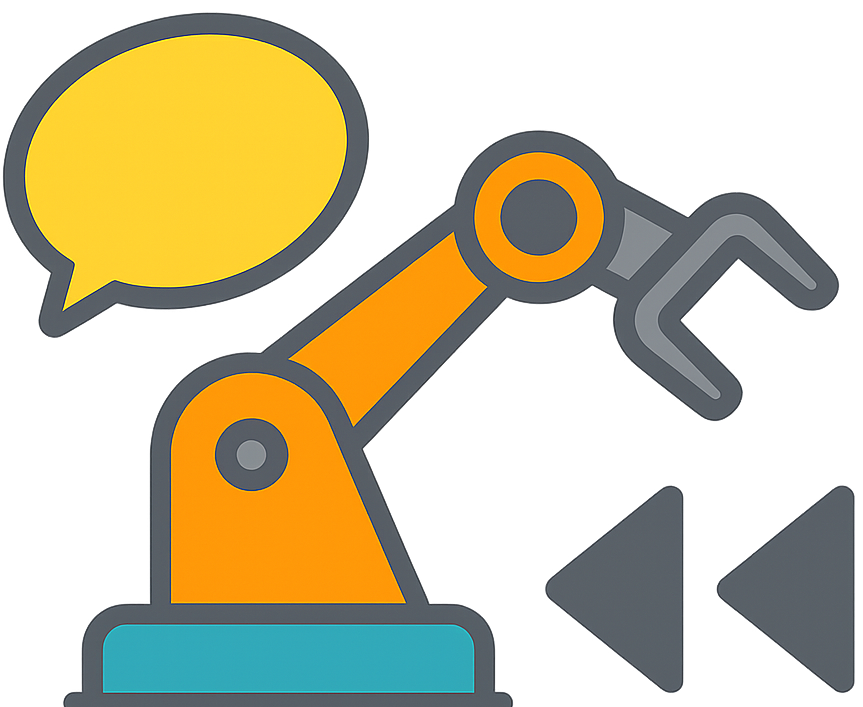}\hspace{0.2em}ReWiND: Language-Guided Rewards\\Teach Robot Policies without New Demonstrations
}

\author{
    Jiahui Zhang$^{\star 1}$, Yusen Luo$^{\star 1}$, Abrar Anwar$^{\star1}$, \\\textbf{Sumedh A. Sontakke$^{2}$},
    \textbf{Joseph J. Lim$^{3}$},
    \textbf{Jesse Thomason$^{1}$, Erdem Bıyık$^{1}$, Jesse Zhang$^{1}$} \\
$^{1}$Thomas Lord Department of Computer Science, University of Southern California\\ 
$^{2}$Amazon Robotics, 
$^{3}$Kim Jaechul School of Artificial Intelligence, KAIST
}

\begin{document}
\maketitle

\begin{abstract}
We introduce ReWiND, a framework for learning robot manipulation tasks solely from language instructions \emph{without per-task demonstrations}. Standard reinforcement learning (RL) and imitation learning methods require expert supervision through human-designed reward functions or demonstrations for every new task. In contrast, ReWiND starts from a small demonstration dataset to learn: (1) a data-efficient, language-conditioned reward function that labels the dataset with rewards, and (2) a language-conditioned policy pre-trained with offline RL using these rewards. Given an unseen task variation, ReWiND fine-tunes the pre-trained policy using the learned reward function, requiring minimal online interaction. We show that ReWiND’s reward model generalizes effectively to unseen tasks, outperforming baselines by up to $2.4\times$ in reward generalization and policy alignment metrics. 
Finally, we demonstrate that ReWiND enables sample-efficient adaptation to new tasks, beating baselines by $2\times$ in simulation and improving real-world pretrained bimanual policies by $5\times$, taking a step towards scalable, real-world robot learning. 
See website at \href{https://rewind-reward.github.io/}{https://rewind-reward.github.io/}.
\end{abstract}

\makeatletter
\if@conferencefinal%
    {\let\thefootnote\relax\footnotetext{$^\star$Equal Contribution }}
\fi
\if@preprinttype%
    {\let\thefootnote\relax\footnotetext{$^\star$Equal Contribution }}
\fi
\makeatother

\section{Introduction}
\label{sec:intro}
A great teacher does not just tell you if you are right or wrong. 
Instead, they guide you by providing feedback when you make mistakes, highlighting progress as you learn something new, and adapting to how you learn best.
For deployed robots to learn new tasks in the wild, they need similarly intelligent teachers. These teachers---in the form of robust reward models---should:
(1) offer \emph{dense, informative feedback}, especially during failures; 
(2) \emph{generalize} their guidance to unseen tasks; and 
(3) remain \emph{robust} to diverse robot behaviors during its learning process.
Our paper leverages these insights to develop reward models capable of teaching robots unseen tasks. 

\begin{figure}[t]
    \centering
    \includegraphics[width=\linewidth]{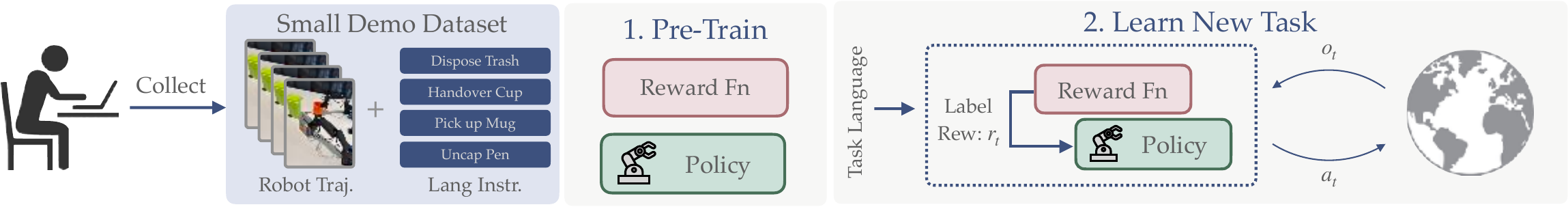}
    \vspace{-0.5cm}
    \caption{\textbf{Overview.} We pre-train a policy and reward model from a small set of language-labeled demos. Then, we solve unseen task variations via language-guided RL \emph{without additional demos}.}
    \label{fig:teaser}
\end{figure}

In this work, we introduce \methodlong, a framework designed to teach robots unseen tasks in a sample-efficient manner using only a few grounding human demonstrations for training tasks (see \Cref{fig:teaser}). 
Typically, teaching robots involves large-scale imitation learning~\citep{brohan2022rt1, brohan2023rt2visionlanguageactionmodelstransfer, black2024pi0visionlanguageactionflowmodel, hamster2025}, where human experts provide demonstrations for each new task. 
However, collecting task-specific demonstrations is expensive and time-consuming. 
Reinforcement learning (RL) offers a more autonomous alternative by using reward functions as teachers, allowing robots to learn through interaction.
Yet, manually designing these reward functions demands substantial manual effort and domain-specific expertise~\citep{rlbook}.
Recent progress in language-conditioned reward learning~\citep{RoboCLIP, kwon2023reward, hu2023language, yu2023language, ma2023eureka, ma2024dreureka, liang2024eurekaverse, alakuijala2024videolanguagecritictransferablereward, wangrlvlmf2024,yang2024trajectory} has aimed at addressing these challenges, but often assumes unrealistic conditions such as availability of ground-truth states~\citep{kwon2023reward, hu2023language, yu2023language, ma2023eureka, ma2024dreureka, liang2024eurekaverse}, thousands of demonstrations~\citep{alakuijala2024videolanguagecritictransferablereward}, or online training of reward models from scratch~\citep{wangrlvlmf2024, yang2024trajectory}, limiting their practical applicability.

\method\ overcomes these challenges by instead assuming only a handful of demonstrations---e.g., five per task---to enable real-world robot learning of unseen task variations.
\method\ first trains a language-conditioned reward model from these demonstrations, then uses it to pre-train a language-conditioned policy via offline RL.
When deployed, \method\ efficiently fine-tunes the policy on new task variations by reward-labeling \emph{online} interaction episodes.

Our core contribution is in designing \method's reward model to capture three key properties outlined earlier: \textbf{dense feedback}, \textbf{generalization}, and \textbf{robustness}.
First, to provide \emph{dense, informative feedback}, we design a \emph{\outputhead} that predicts \emph{progress} within demonstration videos. 
Progress prediction offers a densely supervised training signal that naturally translates into rewards.
Importantly, we also introduce \emph{video rewind} to automatically generate failure trajectories from successful ones, allowing \method\ to provide dense feedback even when the policy is making mistakes.
Then, to encourage \emph{generalization} to unseen tasks and \emph{robustness} to diverse behaviors, we train the sequential aggregator with pre-trained vision and language encoders, selectively applied positional embeddings, and diverse robotics data from Open-X~\citep{open_x_embodiment_rt_x_2023}. Focusing on the above three properties enables \method\ rewards to extrapolate to novel visual and language inputs.

We introduce reward metrics measuring the above properties on which \method\ achieves \textbf{23-74\%} relative improvements over reward learning baselines.
Further, comprehensive success rate evaluations on MetaWorld manipulation tasks and a real-world bimanual robot setup demonstrate \method\ beats baselines by \textbf{2x} in simulation and improves real-world pre-trained policies by \textbf{5x}.

\section{Related Works}
\label{sec:related}

\paragraph{Learning Reward Functions.}
Prior work in reinforcement learning has proposed various methods for \emph{learning} reward functions. Examples include inverse RL~\citep{ng2000IRL, abbeell2004IRL, ziebart2008maxentirl, finn2016gcl}, where reward functions are learned from demonstrations, or methods where rewards are implicitly learned from expert or goal state distributions~\citep{ho2016generative, fu2017learning, fu2018VICE}. 
However, these works require new target-task demonstrations to reward unseen tasks.
\method\ instead trains a general, language-conditioned reward function from an initial demonstration set to reward unseen task variations without further demos.

Another line of work learns reward functions directly from human feedback in the form of comparisons \cite{christiano2017rlhf,sadigh2017active,biyik2020active,lee2021pebble,hejna2022fewshot}, reward sketches \cite{cabi2020scaling}, preference rankings \cite{myers2021learning}, scaled preferences \cite{wilde2021learning}, critiques \cite{cui2018active}, corrections \cite{bajcsy2018learning}, interventions \cite{korkmaz2025mile}, and language \cite{yang2024trajectory}.
While these feedback types may require less human effort than demonstrations or manually written reward functions, these works still require humans to provide extensive feedback for each unseen task.

\paragraph{Reward Generation with Pre-trained Models.}
Prior work has also explored using large pre-trained models to generate reward functions instead of learning them from scratch. Some approaches use LLMs to generate language-conditioned rewards~\citep{kwon2023reward, hu2023language, yu2023language, ma2023eureka, ma2024dreureka, liang2024eurekaverse}, but they typically rely on ground-truth state information that is difficult to obtain in real-world settings. In contrast, \method\ generates rewards from just a task description and a policy execution video.

Other approaches use pre-trained vision models to derive rewards from visual observations~\citep{chen2021learninggeneralizableroboticreward, cui2022can, fan2022minedojo, DECKARD2023, nam2023liftunsupervisedreinforcementlearning, RoboCLIP, wangrlvlmf2024, alakuijala2024videolanguagecritictransferablereward, nguyen2024roboticclipfinetuningclipaction, yang2024rank, yang2023robot, ma2024generative, rocamonde2024visionlanguage, kim2025subtask, hung2025victor}. Among these, RoboCLIP~\citep{RoboCLIP}, LIV~\citep{ma2023eureka}, VLC~\citep{alakuijala2024videolanguagecritictransferablereward}, VICTOR~\citep{hung2025victor}, and GVL~\citep{ma2024dreureka}---like \method---reward unseen manipulation tasks directly from language without sub-task annotations or target-task demos. 
We show in \Cref{sec:experiments:reward_analysis} that these baselines underperform \method\ in rewarding policies in our limited-data setting.
Similar to \method, Foundation Actor-Critic~\citep{ye2024reinforcement} enables efficient RL from language via potential-based shaping rewards from a pre-trained VLM. However, it depends on predefined policy priors (e.g., code-based primitives from LLMs), whereas \method\ learns them through offline RL on non-target tasks.

\section{\method: Learning Rewards Without New Demonstrations}
\label{sec:method}
We study the problem of learning unseen, language-specified tasks in a \emph{target environment}, formulated as a Markov decision process (MDP). The \emph{target environment} refers to the deployment scene (e.g., a robot tabletop). We train a policy $\policy$ that selects actions $a_t$ based on images $o_t$ and language instructions $z$.
The policy is optimized to maximize rewards predicted by a learned reward function $\reward$, which conditions on the frame sequence $o_{1:t}$ and instruction $z$ to output per-timestep estimated rewards $\hat{r}_t$. We assume access to a small demonstration dataset $\demos$ in the target environment containing 15–20 tasks with $\sim$5 demonstrations each. Following prior definitions of generalization~\citep{anwar2024contrast, taxonomy2025arxiv}, we define a task as unseen if it requires a novel \emph{action sequence}, its distribution of \emph{image observations} has changed, or needs a new \emph{language instruction}.

\begin{figure}[t]
    \centering
    \includegraphics[width=\linewidth]{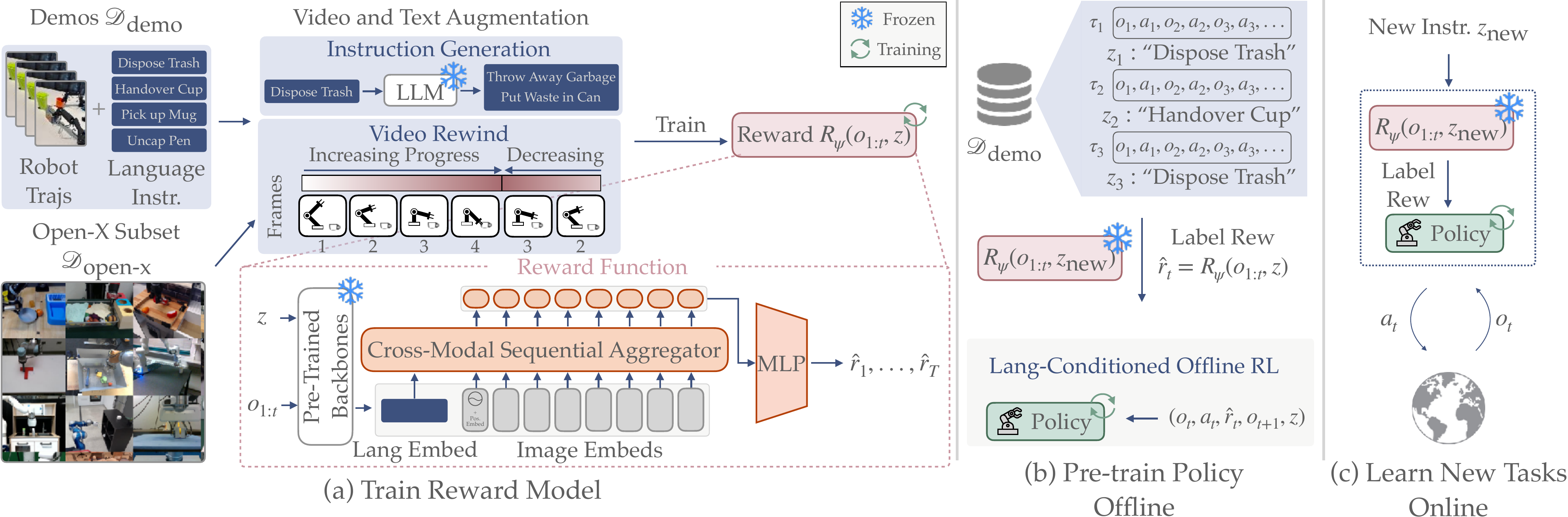}
    \vspace{-0.5cm}
    \caption{
    \textbf{(a):} We train a reward model $\rewardgeneral$ on a small demonstration dataset $\demos$ and a curated subset of Open-X, $\openx$, augmented with LLM-generated instructions and video rewinding. $\reward$ predicts video \emph{progress} rewards $\hat{r}_{1:T}$ from pre-trained embeddings of image observations $o_{1:T}$ and language instructions $z$, and assigns 0 progress to misaligned video-language pairs.
\textbf{(b):} We use the trained $\rewardgeneral$ to label $\demos$ with rewards and pre-train a language-conditioned policy using offline RL.
\textbf{(c):} For an unseen task specified by $z_\text{new}$, we fine-tune the policy with online rollouts and reward labels from $R_\psi(o_{1:t}, z_\text{new})$.
    }
    \label{fig:method}
\end{figure}

\method\ consists of 3 phases (see \Cref{fig:method}): 
(1) learning a reward function from limited target environment demos, then 
(2) pre-training $\pi$ with learned rewards on the demos, and finally
(3) using the reward function and pre-trained policy to learn a new language-specified task online.

\subsection{Learning a Reward Function}
\label{sec:method:reward}
Our primary objective for reward prediction is regressing directly to per-frame \emph{progress} within an observation sequence $o_{1:T}$ conditioned on instruction $z$. Unlike prior methods using relative targets~\citep{yang2024rank, alakuijala2024videolanguagecritictransferablereward}, our progress-based objective provides stable, fixed targets, and translates directly into a dense, $[0, 1]$-normalized reward for policy training. 
To ensure robustness against mismatched observations and instructions, we also sample unrelated observation sequences $o_{1:T}^{\text{other}}$ and train $\reward$ to predict zero progress. Our reward prediction loss is:
\begin{equation}
\resizebox{0.85\linewidth}{!}{%
    $\lossprogress = \sum_{t=1}^T ( \reward - \underbrace{t/T}_{\text{matched seq. progress}} )^2 + \sum_{t=1}^T \underbrace{R_\psi(o_{1:t}^\text{other}, z)^2}_{\text{ mismatched seq. 0 progress}}.$
    }
    \label{eq:progress_loss}
\end{equation}
However, simply training a neural network $\rewardgeneral$ on $\lossprogress$ with a small set of demonstrations is unlikely to ensure that it can train a policy on unseen tasks.
$\rewardgeneral$ should:

\begin{enumerate}[label=\textbf{D\arabic*}, nosep] %
    \item\label{dgeneralize} \textbf{Generalize to new tasks}, i.e., new policy execution videos and instructions not in $\demos$.
    \item\label{dreward} \textbf{Produce rewards aligned with \emph{policy rollouts}}, not just successful demonstration videos.
    \item\label{drobust} \textbf{Be robust to input variations}, i.e., different ways to solve or specify the task.
\end{enumerate}

\newcommand{\dgeneralize}{\ref{dgeneralize}}
\newcommand{\dreward}{\ref{dreward}}
\newcommand{\drobust}{\ref{drobust}}

\newtext{To this end, we introduce a set of design choices spanning the training dataset, model architecture, and video and language augmentations that address all three desiderata.} Specifically, we curate diverse off-the-shelf data from the Open-X dataset~\citep{open_x_embodiment_rt_x_2023} to promote generalization (\dgeneralize) and robustness (\drobust); apply targeted video and language augmentations for better reward prediction and language input robustness (\dreward, \drobust); and adopt specific network architectural modifications aimed at improving generalization (\dgeneralize).
For a visual overview, see \Cref{fig:method}a.

\textbf{Incorporating Diverse Data (\dgeneralize, \drobust).}
To help $\rewardgeneral$ generalize to tasks unseen in $\demos$ (\dgeneralize) and make it robust to diverse ways of executing and specifying tasks (\drobust), we subsample the Open-X Dataset~\citep{open_x_embodiment_rt_x_2023}, denoted $\openx$.
We specifically select Open-X trajectories with object-centric language instructions, e.g., ``\textit{pick coke can from fridge},'' or directional instructions, e.g., ``\textit{drag the circle to the left of the star},'' to help $\rewardgeneral$ generalize to objects and directions not contained in $\demos$.
This dataset contains $\sim$356k trajectories with $\sim$59k unique task strings. 
For detailed dataset information, see \Cref{sec:appendix:dataset}.

\subsubsection{Video and Language Augmentation (\dreward, \drobust)}
\label{sec:method:reward:augmentation}
Given our datasets $\demos$ and $\openx$, we perform both video and language augmentations that help the reward function accurately predict rewards for unsuccessful \emph{policy execution} videos (\dreward) and be robust to varied ways of specifying the task instructions $z$ (\drobust). 
We call the video augmentation \textbf{video rewind} and our text augmentation \textbf{instruction generation}.

\begin{wrapfigure}[13]{r}{0.46\textwidth}
    \centering
    \vspace{-0.3cm}
    \includegraphics[width=\linewidth]{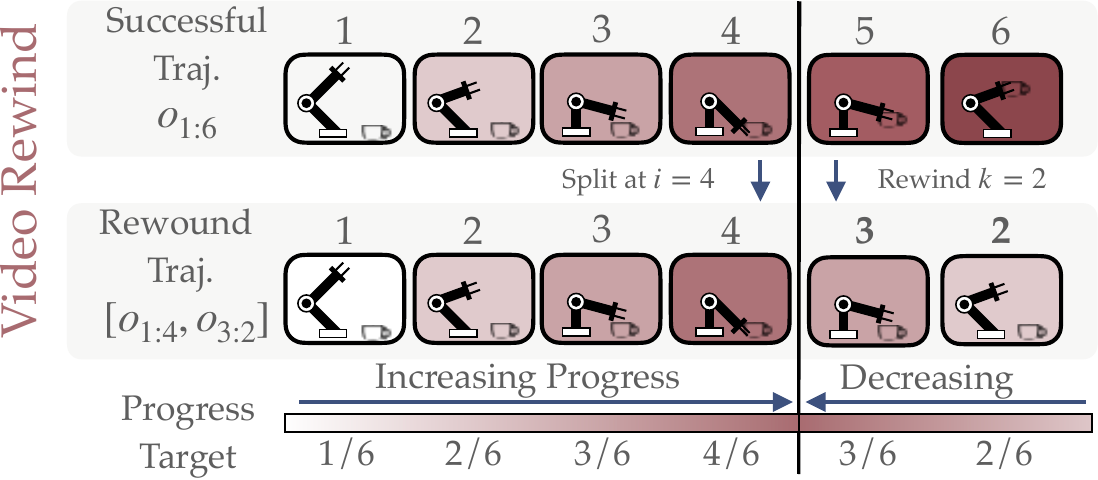}
    \vspace{-0.4cm}
    \caption{
    \textbf{Video rewind.} 
    We split a demo at intermediate timestep $i$ into forward/reverse sections. Here, the forward section shows the robot approaching the cup; the reverse section ($o_{i-1}, o_{i-2}, \ldots$) resembles dropping it. 
    }
    \label{fig:video_rewind}
\end{wrapfigure}
\textbf{Video Rewind.} Both $\demos$ and $\openx$ contain human demonstrations, which are assumed to be successful and of high-quality.
Training $\rewardgeneral$ on $\lossprogress$ only using these successful demonstrations, may result in $\rewardgeneral$ overfitting to these successful trajectories.  
However, during online deployment, $\rewardgeneral$ will likely encounter failure trajectories (unseen during training) which such an overfit model may reward highly. 
This is undesirable and prior works attempt to address this issue by explicitly training their reward model on failed trajectories~\citep{alakuijala2024videolanguagecritictransferablereward}, but these trajectories add a great additional burden on demonstrators to collect and must be added post-hoc to any existing dataset, making it harder to scale.

Instead, we address this problem in a scalable manner by randomly \emph{rewinding videos}. 
Consider a video of a robot picking up a cup. 
If we rewind the video for a few frames right when the robot grabs the cup, it now looks like one in which the robot attempted to grasp the cup and then dropped it.\footnote{Random rewinding may result in some physically implausible sequences. However, since they won't appear during inference, the rewards produced by $\rewardgeneral$ for such sequences should not affect online RL.}
By training $\rewardgeneral$ to predict rewards corresponding to \emph{reverse progress} on the rewound subsequence, it (1) is trained on observation sequences mimicking failed policy rollouts that will occur during online RL, and (2) learns to \emph{decrease} reward when necessary. 
Thus rewinding helps $\rewardgeneral$ reward a \emph{policy's} \newtext{failures }which will help with online RL (\dreward).
See \Cref{fig:video_rewind} for a visual example.
Formally, rewinding means sampling a random split point $i$ within an observation sequence $o_1...o_T$, rewinding $k$ ($k$ is also sampled) frames, then concatenating those $k$ frames to the end of the original sequence to become $o_1...o_i, o_{i-1},...,o_{i-k}$.
The remaining frames from $i+1$ to $T$ are then unused.
Our video rewind training objective follows:
\begin{equation}
    \lossrewind =  \sum_{t=1}^{i} \underbrace{( \reward  - \frac{t}{T} )^2}_{\text{Loss for original trajectory until } i}  + \sum_{t=1}^k \underbrace{( R_\psi([o_{1:i}, o_{i-1:i-t}], z)- \frac{i - t}{T} )^2}_{\text{Rewound video for } k \text{ frames from } i - 1}.
    \label{eq:rewind}
\end{equation}

\textbf{Instruction Generation.}
We also generate 5-10 additional language instructions for each task in $\demos$ by prompting an LLM.
This augmentation helps $\reward$ with input robustness to possible new task instructions (\drobust). 
While training $\reward$, any time we sample an observation sequence $o_{1:T}$, its instruction $z$ is uniformly randomly sampled from all available matching instructions, generated or original. 
We did not augment $\openx$ due to its instruction diversity.

\subsubsection{Architecture (\dgeneralize)}
\label{sec:method:reward:arch_and_objective}
Due to the limited size of $\demos$, we carefully design the architecture for $\rewardgeneral$ to maximize generalization to new tasks (\dgeneralize) while retaining the ability to optimize $\lossprogress$ well. 

\textbf{Frozen Input Encoders.} We use frozen image and language encoders as the backbone of $\rewardgeneral$: we use DINOv2~\citep{oquab2024dinov2learningrobustvisual} for image encoding due to its strong object-centric representations and \texttt{all-MiniLM-L12-v2}~\citep{reimers-2019-sentence-bert} for instruction encoding due to its small embedding size ($=384$).
In $\rewardgeneral$, we first encode images and instructions: $o_{1:t}^\text{embed} = \text{DINO}(o_{1:t}), z^\text{embed} = \text{MiniLM}(z)$.
Then, we train a small \emph{\outputhead} transformer conditioned on $(o_{1:t}^\text{embed}, z^\text{embed})$ that learns to aggregate frozen language and image embeddings to generate progress rewards $\hat{r}_t$ directly (see \Cref{fig:method}(a) in the ``Reward Function'' box).

\textbf{Positional Embeddings.} Finally, the \outputhead's transformer requires positional information about the frames to properly predict rewards (e.g., for distinguishing ``pull'' vs. ``push'').
However, if we na\"{i}vely add positional embeddings to each image, it can ``cheat'' by predicting progress using the positional embeddings.
Therefore, similar to how \citet{ma2024generative} prompt an LLM with the position of the first video frame, we add a positional embedding to the \emph{first} image.

\textbf{Reward Model Summary.}
In summary, \method\ trains a reward function $\rewardgeneral$ to predict task progress, using data augmentation (video rewinding \newtext{and} instruction generation) and additional Open-X data ($\openx$) to improve generalization. $\rewardgeneral$ combines pretrained vision and language encoders with a lightweight \outputhead\ that uses only first-frame positional embeddings. For full implementation details, see \Cref{sec:appendix:implementation:rewind_implementations:reward}. The final objective is:
\begin{equation}
\text{min}_\psi \; \mathbb{E}_{(o_{1:T}, z, o^\text{other}_{1:T}) \sim {\demos, \openx}} \big[ \lossprogress + \lossrewind \big].
\end{equation}

\subsection{Policy Learning}
\label{sec:method:policy learning}
\textbf{Pre-training.} After training $\rewardgeneral$, we pre-train $\policy$ on demonstrations $\demos$ labeled with rewards. This pre-training guides $\policy$ toward reasonable behaviors during exploration, even if downstream tasks differ from those in $\demos$.
Given a trajectory with instruction $z$, $\{(o_t, a_t)\}_1^T$, we assign rewards $\hat{r}_t = \reward$ at each timestep and add a success bonus to the final reward to encourage reaching the goal despite possibly noisy reward signals:
\begin{equation}
\offlinerewards_t =
\reward + \success \cdot \mathbbm{1}[t = T].
\label{eq:offline_labeled_rewards}
\end{equation}

We then train $\policy$ via offline RL using tuples $(o_t, a_t, \hat{r}_t, o_{t+1}, z)$. We use IQL~\citep{kostrikov2022offline} as prior work has demonstrated it works on real robots~\citep{venkataraman2024real,zhang2023bootstrap, zhang2024sprint}. See \Cref{fig:method}(b) for an overview.

\textbf{Learning Online.}
To learn a new task online, \method\ only requires a language description of the task, $z_\text{new}$.
\method\ rolls out $\pi(a \mid o_t, z_\text{new})$ and fine-tunes it on rewards coming from $R_\psi(o_{1:t}, z_\text{new})$.
Like prior work~\citep{alakuijala2024videolanguagecritictransferablereward, yang2024rank}, we assume access to a success signal during online RL. We use this signal to give $\success$ bonuses similar to pre-training.\footnote{Success bonuses can come from a human supervisor~\citep{luo2024hilserl}, learned function~\citep{fu2018VICE}, or LLM~\citep{ye2024reinforcement}. Our experiments assume a human supervisor because manual resets are required regardless. 
While we could threshold $\reward$ outputs to automatically determine success, unseen evaluation task reward ranges can vary, rendering this approach ineffective.
Future work could integrate \method\ with methods reducing human resets~\citep{yang2023robot, gupta2021reset} and automatic success detectors for truly autonomous RL.
}
Our online rewards $\onlinerewards$ are:
\begin{equation}
\label{eq:online_rewards}
    \onlinerewards_t = \reward + \success \cdot \mathbbm{1}[\text{success at } t].
\end{equation}
See full implementation details in \Cref{sec:appendix:implementation:rewind_iplmeentations} and pseudocode in \Cref{alg:full_approach_overview}.

\section{Experiments}
\label{sec:experiments}

Our experiments aim to study the efficacy of \method\ as a reward learning pipeline, evaluate its ability to train robots to learn new tasks efficiently, and analyze its design choices and limitations. To this end, we organize our experiments to answer the following empirical questions, in order:
\begin{enumerate}[label=(\textbf{Q\arabic*}), nosep]
    \item \label{q1} \textbf{Rewards}: How well do \method\ rewards correlate with task progress and success?
    \item \label{q2} \textbf{Policy Learning}: Can \method\ quickly train policies for new tasks?
    \item \label{q3} \textbf{Ablations and Analysis}: Which \method\ design decisions are most significant?
\end{enumerate}

\subsection{Q1: What Makes a Good Reward Function?}
\label{sec:experiments:reward_analysis}
We repeat the desiderata from \Cref{sec:method:reward} that we set out to achieve with \method: (1) generalization to new tasks, (2) rewards aligned with videos from \emph{policy rollouts}, and (3) robustness to diverse inputs.
We structure this section to demonstrate \method's ability to satisfy these criteria.

We compare \method-learned rewards against all relevant reward learning baselines from \Cref{sec:related}:
\textbf{LIV}~\citep{ma2023liv} is a robotics reward model pre-trained on EpicKitchens~\citep{Damen2022EpicKitchens}, we also fine-tune LIV on $\demos$ (\textbf{LIV-FT});
\textbf{RoboCLIP}~\citep{RoboCLIP} uses a pre-trained video language model, S3D~\citep{xie2018rethinking} trained on HowTo100M~\citep{miech2019howto100m}, to reward agents for language specified tasks; 
\textbf{Video-Language Critic (VLC)}~\citep{alakuijala2024videolanguagecritictransferablereward} fine-tunes a VLM with a sequential ranking objective to encourage frames later in the video to have higher rewards. We train it on $\demos$;
\textbf{Generative Value Learning (GVL)}~\citep{ma2024generative} prompts a pre-trained Gemini LLM~\citep{geminiteam2024geminifamilyhighlycapable} with shuffled frames to predict per-frame progress.

We conduct our primary reward analysis using the simulated MetaWorld benchmark~\citep{yu2021metaworldbenchmarkevaluationmultitask} because it enables efficient collection of exemplar failed and partially successful rollout videos for analysis. Smaller-scale real-world analyses, \textbf{strongly aligned} with simulation, are in \Cref{sec:appendix:additional_results:real_world_reward}. 
$\demos$ here consists of 20 tasks with 5 expert demos each.
For fair comparison, we include a variant of \method\ trained without $\openx$ (\textbf{\method\ w/o OXE}). Results are evaluated on 17 unseen MetaWorld tasks. 
\newtext{These tasks are visually similar to training tasks but require new motions to solve (e.g., \texttt{Door-Open} $\rightarrow$ \texttt{Door-Close}).}
We average metrics across 5 demos per task.

\textbf{Generalization.} 
\begin{figure}[t]
    \centering
    \includegraphics[width=\linewidth]{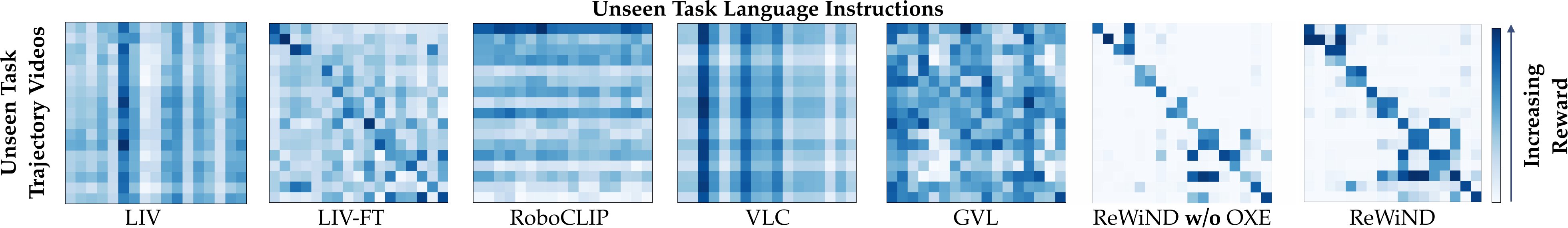}
    \vspace{-0.4cm}
        \caption{\textbf{Video-Language Reward Confusion Matrix.} 
        For each unseen \newtext{MetaWorld} task, we compute rewards for all combinations of demonstration videos and language descriptions. 
        \method\ produces the most \textcolor{blue}{diagonal-heavy} confusion matrix, indicating strong alignment between unseen demos and instructions. See \Cref{sec:appendix:additional_results:reward_analysis} for train task results, \newtext{\Cref{sec:appendix:additional_results:real_world_reward} for real-world results.}
        }
    \label{fig:exps:confusion_matrices}
\end{figure}
We first evaluate how effectively each reward model distinguishes unseen tasks using confusion matrices of unseen task videos versus language instructions (\Cref{fig:exps:confusion_matrices}). Ideally, a clear \textcolor{blue}{blue} diagonal indicates correct video-instruction pairs, with low (white) values elsewhere. \method\ produces clearest disparity between the diagonal and off-diagonal elements, excelling even without OXE due to architectural choices aimed at generalization, i.e., first frame positional encodings and frozen pre-trained input embeddings. 

\begin{table}[t]
\centering
\caption{\textbf{Combined Evaluation Metrics.} Comparison of reward models across three axes: (1) Demo Video Reward Alignment, (2) Policy Rollout Reward Ranking, and (3) Input Robustness. }
\label{tab:combined_results}
\resizebox{\linewidth}{!}{%
\begin{tabular}{@{}llccccccc@{}}
\toprule
\textbf{Category} & \textbf{Metric} & \textbf{LIV} & \textbf{LIV-FT} & \textbf{RoboCLIP} & \textbf{VLC} & \textbf{GVL} & \textbf{\method{} w/o OXE} & \textbf{\method{} w/ OXE} \\
\midrule

\multirow{2}{*}{\textbf{(a) Demo Reward Alignment}} 
    & $r$ $\uparrow$       & -0.03 & 0.55 & 0.01  & 0.64 & 0.52 & 0.67 & \textbf{0.83} \\
    & $\rho$ $\uparrow$    & -0.04 & 0.55 & -0.01 & 0.62 & 0.57 & 0.64 & \textbf{0.79} \\

\midrule

\multirow{2}{*}{\textbf{(b) Policy Rollout Ranking}} 
    & Rew. Order $\rho$ $\uparrow$ & -0.32 & 0.47 & 0.00 & -0.18 & 0.32 & 0.76 & \textbf{0.82} \\
    & Rew. Diff. $\uparrow$ & -0.16 & 0.26 & 0.06 & -0.15 & 0.17 & 0.39 & \textbf{0.41}\\

\midrule

\multirow{2}{*}{\textbf{(c) Input Robustness}} 
    & Avg. $\rho$ $\uparrow$        & 0.03 & 0.27 & 0.00 & 0.60 & 0.58 & 0.55 & \textbf{0.74} \\
    & $\rho$ Variance $\downarrow$ & 0.08 & 0.28 & \textbf{0.00} & \textbf{0.00} & 0.01 & 0.03 & 0.04 \\

\bottomrule
\end{tabular}
}
\end{table}

Next, we evaluate how consistently rewards reflect progress over time in successful, unseen demonstrations. We report Pearson correlation ($r$) of each model's per-frame reward against time and also Spearman’s rank correlation ($\rho$), which, unlike $r$, captures monotonicity regardless of linearity.
As shown in \Cref{tab:combined_results}\textbf{(a)}, \method\ again outperforms all baselines—achieving a \textbf{30\%} relative improvement in $r$ and \textbf{27\%} in $\rho$ over the best alternative (VLC).

\textbf{Policy Rollout Reward Alignment.}
We also find that \method\ can properly reward \emph{failed} policy rollouts, which is important for rewarding RL policies on unseen tasks.
For each task, we train an SAC~\citep{haarnoja2017soft} policy from scratch and use trajectories collected from various points of training to construct three evaluation video datasets: \texttt{failure}, \texttt{near-success}, and \texttt{success} containing failed trajectories, trajectories where the policy was close to the goal state but did not succeed, and successful trajectories, respectively. Each task has 2 trajectories of each type.

We evaluate each dataset's relative alignment \emph{ranking} (measured by Spearman's $\rho$) with each reward model. For example, for a given task, if the average reward for a \texttt{failure} video is 0.1, a \texttt{near-success} video is 0.5, and \texttt{success} video is 0.9, then the rankings would be 1, 2, 3, respectively, where 3 corresponds to the best ranking.
Thus, $\rho$ over rankings tells us how often the videos are correctly ranked.
We report the ranking in \Cref{tab:combined_results}\textbf{(b)}.
We also report the average \emph{difference} between rewards for \texttt{success} with \texttt{near-success} and \texttt{near-success} with \texttt{failure} videos.
Overall, due to \emph{video rewind}, \method\ has a relative \textbf{74\%} improvement in reward order and \textbf{58\%} improvement in reward difference over the best baseline, LIV-FT. 
Additionally, we qualitatively demonstrate how these rankings translate into policy rollout rewards in Appendix \Cref{fig:exps:policy_perf_analysis} by plotting reward predictions of \method\ against reward baselines for an unsuccessful policy rollout.

\textbf{Robustness to Varied Inputs.}
Finally, we demonstrate \method's robustness to diverse instructions.
For each evaluation task, we manually create three additional language instructions (without prior knowledge of \method’s performance), resulting in four total instructions per task. For example, ``close the door'' is an original instruction, and we add ``shut the door.'' Each set of instructions is paired with a single demonstration video, and we compare the reward models by measuring their average Spearman’s rank correlation ($\rho$) and output variance across these instructions in \Cref{tab:combined_results}\textbf{(c)}. Higher variance indicates lower robustnes.
Again, \method\ outperforms baselines, achieving the highest average correlation ($0.74$), \textbf{23\%} better than VLC, and near-zero variance, even without OXE training---likely aided by our instruction augmentation approach (\Cref{sec:method:reward:augmentation}). RoboCLIP and VLC show near-zero variance but achieve significantly lower correlation scores.

So far, our results demonstrate that \textbf{\method\ significantly outperforms all image-language-conditioned reward baselines} in terms of \textbf{generalization}, rewarding \textbf{policy rollouts}, and input \textbf{robustness}.
We next demonstrate how these results translate into sample-efficient policy learning.

\subsection{Q2: Learning New Tasks with RL}
\label{sec:experiments:policy}
\begin{wrapfigure}[12]{r}{0.4\textwidth}
    \vspace{-0.5cm}
    \centering
    \includegraphics[width=\linewidth]{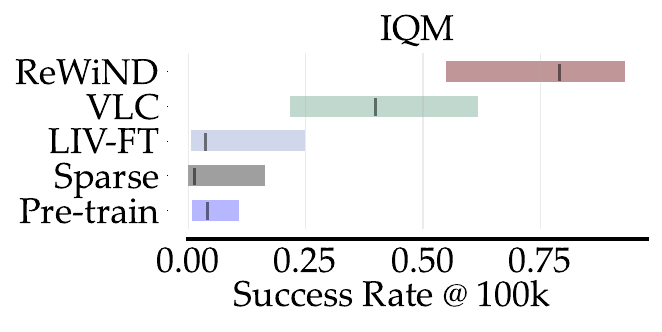}
    \caption{
    \textbf{MetaWorld final performance.} We plot inter-quartile means (IQMs) of success rates after 100k environment steps on 8 unseen tasks in MetaWorld. \method\ achieves 79\%.
    }
    \label{fig:exps:metaworld_results}
\end{wrapfigure}

\textbf{Simulation.}
We use the MetaWorld simulation benchmark~\citep{yu2021metaworldbenchmarkevaluationmultitask}, where we pre-train reward models and policies on 20 tasks, each with 5 per-task demos collected from a scripted policy. 
We evaluate on 8 unseen tasks in MetaWorld, chosen for reasonable initial policy rollout behaviors, across 3 seeds each. 
We compare \method\ against the 2 language-conditioned reward model baselines that performed best in reward alignment (\textbf{VLC}) and policy rollout rankings (\textbf{LIV-FT}) from the reward analysis in  \Cref{sec:experiments:reward_analysis}.
We also compare against \textbf{Sparse}, which pre-trains and fine-tunes on only the sparse success reward bonus, and \textbf{Pre-train}, which pre-trains on sparse reward and is evaluated zero-shot on new tasks.
All baselines are image, proprio ($x, y, z, \text{gripper}$), and language conditioned.
Each method uses the same pre-training and RL procedure as \method\ (\Cref{sec:method:policy learning}), and is trained online for 100k timesteps. 
See Appendix~\ref{sec:appendix:metaworld} for more experiment details.

As recommended by \citet{agarwal2021deep}, we report the interquartile mean (IQM) and 95\% confidence intervals computed over all task success rates at 100k environment steps in \Cref{fig:exps:metaworld_results}. 
Sparse reward fine-tuning and Pre-train (no fine-tuning) result in near-zero success rates, highlighting the difficulty of image-based new task learning under limited data. 
In fact, Sparse reward fine-tuning, which relies purely on a sparse success bonus, performs worse than Pre-train after fine-tuning.
Meanwhile, \method\ achieves an IQM success rate of \textbf{79\%}, a \textbf{97.5\%} improvement over the best baseline, VLC, demonstrating that \method\ effectively enables the policy to learn new tasks in MetaWorld.
These results are well-aligned with our reward analysis in \Cref{sec:experiments:reward_analysis}, demonstrating how they correlate with policy learning performance.
\method\ is also more sample-efficient at timesteps less than 100k; see extended discussion in \Cref{sec:appendix:addtional_results:metaworld_sample_efficiency} and sample efficiency curves in \Cref{fig:metaworld view}.

\textbf{Real-World Robot Learning.}
\begin{figure}[t]
    \centering
    \includegraphics[width=\textwidth]{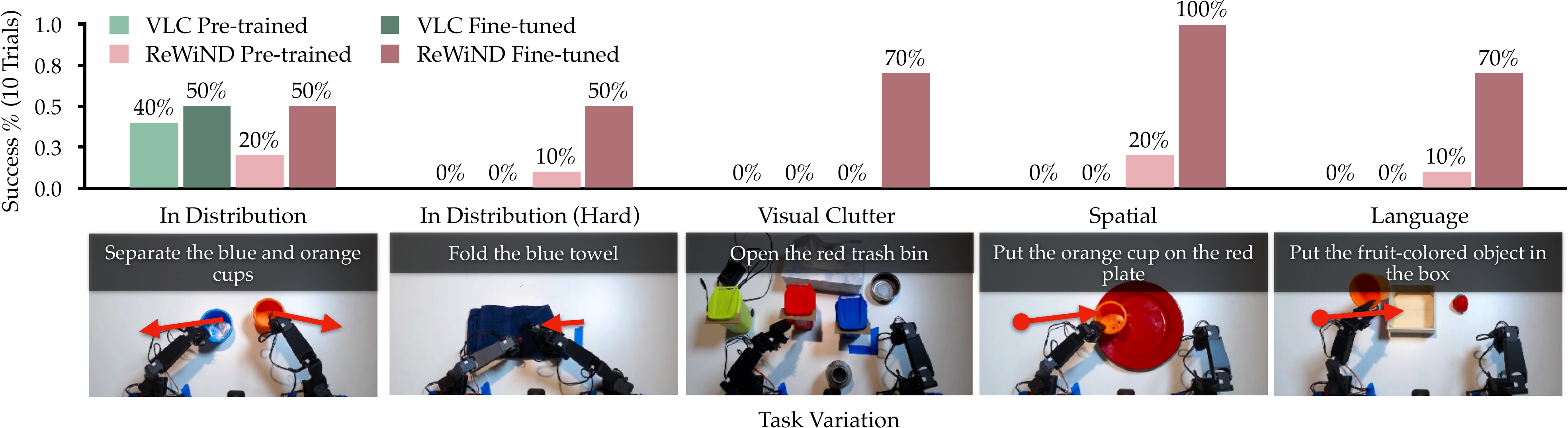}
    \caption{\textbf{Real-robot RL.}
    We present results on the Koch bimanual arms across in-distribution tasks and visual, spatial, and linguistic generalization tasks.
    Online RL with \method\ improves a pre-trained policy by an absolute 56\% across all five tasks.
    }
    \vspace{-0.5cm}
    \label{fig:exps:real_world_results}
\end{figure}
We conduct real-world tabletop manipulation experiments with a bimanual Koch v1.1 robot arm setup~\citep{cadene2024lerobot}.
\newtext{We use 5 demos to train the reward function, but 10 for the policy, as we found policy learning to be a bottleneck on this difficult robot embodiment}.
Across five tasks, we demonstrate in \Cref{fig:exps:real_world_results} that an hour of real-world RL with \method\ improves the success rate over the base pre-trained policy from an average 12\% success rate to 68\%, a \textbf{5$\times$} improvement. 
\newtext{Meanwhile, VLC only improves from 8\% to 10\%---\method\ outperforms VLC, the best simulation baseline, by \textbf{6.7$\times$}.}
RL for an hour of real-world experiment time corresponds to 50k environment steps with our parallelized codebase that trains the policy while an older checkpoint gathers data in the environment to avoid any training wait time. 
We select diverse tasks that demonstrate real-world improvement based on generalization metrics defined in prior work~\citep{anwar2024contrast, taxonomy2025arxiv} on: an in-distribution task, \texttt{separate the blue and orange cups}; an in-distribution \emph{difficult} task, \texttt{fold the blue towel}; an unseen task in terms of large amounts of \emph{visual} clutter, \texttt{open the red trash bin}; an unseen task in terms of spatial relationships between objects requiring \emph{new action sequences}, \texttt{put the orange cup on the red plate}; and an unseen task in terms of \emph{language} input, \texttt{put the fruit-colored object in the box}.
Overall, \method\ enables real-world reinforcement learning on unseen tasks without requiring new demonstrations, improving over the pre-trained policy, and outperforms the best baseline from simulation, VLC.
See \Cref{sec:appendix:real_robot:setup} for real-world experiment details and \Cref{fig:exps:rollouts} for policy rollout examples.

\begin{table}[t]
\centering
\caption{\textbf{Ablation Study:} subtracting ($-$) and adding $+$ various \method\ components on \newtext{MetaWorld} training and evaluation task (a) demo reward alignment, evaluation task (b) policy rollout ranking order and reward difference, and evaluation task (c) input robustness.}
\label{tab:ablations}
\resizebox{\textwidth}{!}{%
\begin{tabular}{@{}lcccccc@{}}
\toprule
\multirow{2}{*}{Model} & \multicolumn{2}{l}{(a) Demo Reward Alignment}   & \multicolumn{2}{l}{(b) Policy Rollout Ranking}  & \multicolumn{2}{l}{(c) Input Robustness}        \\ \cmidrule(l){2-7}
 & Train Demos $\rho$ $\uparrow$ & Unseen Demo $\rho$ $\uparrow$ & Rew. Order $\rho$ $\uparrow$ & Rew. Diff. $\uparrow$ & Avg. $\rho$ $\uparrow$ & $\rho$ Variance $\downarrow$ \\ \midrule
\textbf{Original \method}  & \textbf{1.00} & 0.79 & \textbf{0.82} & \textbf{0.41} & 0.74 & 0.04 \\ \midrule
 \textbf{$-$ Targ. Env Data} &  0.55 &  0.77 &  0.18 & 0.08  &  \textbf{0.78}  & 0.04  \\
 \textbf{$-$ Open-X Subset} &  \textbf{1.00} & 0.64  & 0.76  & 0.39  & 0.55  & 0.03  \\
 \textbf{$-$ Video Rewind} & \textbf{1.00}  & 0.69  & 0.56  & 0.27  & 0.66  &  \textbf{0.02} \\
 \textbf{$-$ Instr. Generation}& \textbf{1.00}  & 0.66  &  0.62 & 0.30  & 0.52  & 0.07  \\
 \textbf{$+$ Full Pos. Embeds}& 0.99  & \textbf{0.85}  & 0.71  & 0.33  & \textbf{0.78}  & 0.06 \\ \bottomrule
\end{tabular}
}
\end{table}
\begin{table}[t]
\centering
\caption{\textbf{Ablation Policy Performance results}: Select reward metrics and policy success rates for ablations, normalized out of 1.00 by ReWiND's original performance.}
\label{tab:rewind_variations_norm}
\resizebox{0.8\linewidth}{!}{%
\begin{tabular}{@{}lcc|c@{}}
\toprule
\textbf{ReWiND Variant} &
\textbf{Rank Order Corr. $\uparrow$} &
\textbf{Avg.\ Robustness Corr.\ $\uparrow$} &
\textbf{Policy Succ.\ Rate $\uparrow$} \\ 
\midrule
5\,Task & 0.76 & 0.86 & 0.52 \\
10\,Task & 0.90 & 0.88 & 0.47 \\
\midrule
+ Full\,PE     & 0.71 & \textbf{1.05} & 0.69 \\
- Video Rewind   & 0.87 & 0.89 & 0.67 \\
ReWiND  & \textbf{1.00} & 1.00 & \textbf{1.00} \\
\bottomrule
\end{tabular}}
\end{table}

\subsection{Q3: Ablation Study} 
We perform a thorough ablation study of \method\ regarding how specific design choices influence demonstration reward alignment, policy rollout ranking, and input robustness metrics introduced in \Cref{sec:experiments:reward_analysis}.
We ablate: instruction generation and video rewinding (\Cref{sec:method:reward:augmentation}); using OXE data; the need for target environment data $\demos$; and finally, the use of first frame vs. full frame positional embeddings on the input observation sequence $o_{1:T}$ in the \outputhead\ (\Cref{sec:method:reward:arch_and_objective}).
Overall, the original \method\ model performs best across most metrics. Below, we analyze the impact of each ablation:

\textbf{Datasets.}
Removing target environment data (\textbf{$-$Targ. Env Data})—i.e., using only $\openx$ data without $\demos$—leads to poor alignment with training demonstrations (\Cref{tab:ablations}a) and fails to distinguish between failed, near-successful, and successful policy rollouts (\Cref{tab:ablations}b). However, it retains strong input robustness due to the diversity of OXE data.
Meanwhile, removing the Open-X subset (\textbf{$-$Open-X Subset}) harms unseen task reward alignment (\Cref{tab:ablations}a) and input robustness (\Cref{tab:ablations}c), highlighting the importance of OXE data for generalizing across varied language instructions.

\textbf{Augmentation.} Eliminating video rewinding (\textbf{$-$Video Rewind}) degrades rollout ranking performance (\Cref{tab:ablations}b), showing that rewinding helps distinguish failed rollouts as intended. This variant performs similarly to the single-image LIV-FT baseline in Table~\ref{tab:combined_results}, indicating that video rewinding more effectively captures the temporal information in the videos. \cameraready{It also reduces policy learning performance by 33\% as seen in \Cref{tab:rewind_variations_norm}}.
Similarly, removing instruction generation (\textbf{$-$Instruction Generation}) reduces performance on language input robustness (\Cref{tab:ablations}c), confirming that LLM-generated instructions enhance robustness to diverse inputs.

\textbf{Architecture.} Adding full positional embeddings (\textbf{$+$Full Pos. Embeds}) improves unseen demo alignment (\Cref{tab:ablations}a) but worsens rollout ranking (\Cref{tab:ablations}b), likely due to overfitting---where the model learns to predict increasing rewards regardless of input. \cameraready{This overfitting results in 21\% inferior relative policy success rate, as seen in \Cref{tab:rewind_variations_norm}.} To avoid overfitting, the main \method\ model uses only first-frame positional embeddings (\Cref{sec:method:reward:arch_and_objective}).

\textbf{Less Training Tasks.}
\cameraready{Finally, we additionally check the performance of \method\ with less training tasks: 5 and 10 instead of the original 20. 
We can see in \Cref{tab:rewind_variations_norm} that the reward model still performs well even with just 5 tasks of training data instead of 20, but policy success rates understandably drop with fewer tasks.}

\begin{figure}[ht]
   \includegraphics[width=\linewidth]{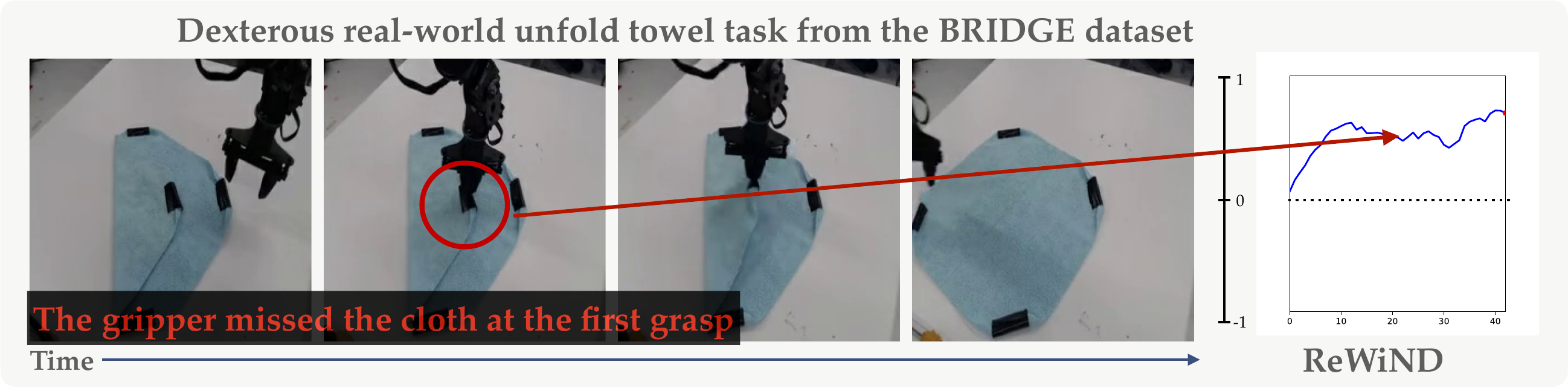}
   \caption{ReWiND reward for a contact-rich BRIDGE task. Rewards stagnate when the gripper misses, then increase as the task succeeds.}
   \label{fig:dexterous fig}
   \vspace{-1.5em}
\end{figure}

\subsection{Additional Analysis and Results}
\paragraph{Contact Rich BRIDGE Example.} \cameraready{We additionally plot an example of a \method's predictions for a contact-rich unfolding towel task from the BRIDGE-v2 dataset~\citep{walke2023bridgedata} contained in the OXE~\citep{open_x_embodiment_rt_x_2023} training data used to train \method in \Cref{fig:dexterous fig}.
Even though the \method\ reward function was not trained for many training steps overall in order to prevent overfitting (2k in MetaWorld and 10K in the Real-World, see \Cref{sec:appendix:implementation:rewind_implementations:reward}), it can perform well on this BRIDGE example. Notice specifically that \method\ rewards stagnate when the gripper first misses the cloth grasp, and then later increase again as it succeeds on the second retry.
}

\paragraph{Diffusion Steering.} \cameraready{We conducted preliminary experiments with ReWiND on more expressive policy classes, namely diffusion policies (DPs)~\cite{chi2023diffusion}.
We use Diffusion Steering RL (DSRL)~\citep{wagenmaker2025steering} for online RL of DPs.
Rather than optimizing the weights of the DP, DSRL runs RL over the latent-noise space of a diffusion policy.
This latent space encodes an action prior that RL can refine to improve DPs. The RL policy, conditioned on image and instruction, was trained offline with ReWiND rewards and finetuned online in MetaWorld (Appendix~\ref{sec:appendix:metaworld:training}).
For seen tasks, this approach improved base DPs: offline RL learned useful priors that accelerated online RL.
However, when we attempted to learn \textit{new} tasks, we found that the action priors learned from offline RL was not representative of the new task as the instruction (OOD and data-sparse regime).
We thus trained the latent policy only on images (while keeping the DP instruction-conditioned)
As a preliminary result, this DSRL+ReWiND implemented improved DP's performance on an unseen task \textbf{0\% $\rightarrow$ 20\%}.
As such, ReWiND is agnostic to policy architectures: single-action-output MLP architectures in MetaWorld, action-chunked transformers in the real world, and now also DPs via diffusion steering.
}

\begin{figure}[t]
    \centering
    \includegraphics[width=\textwidth]{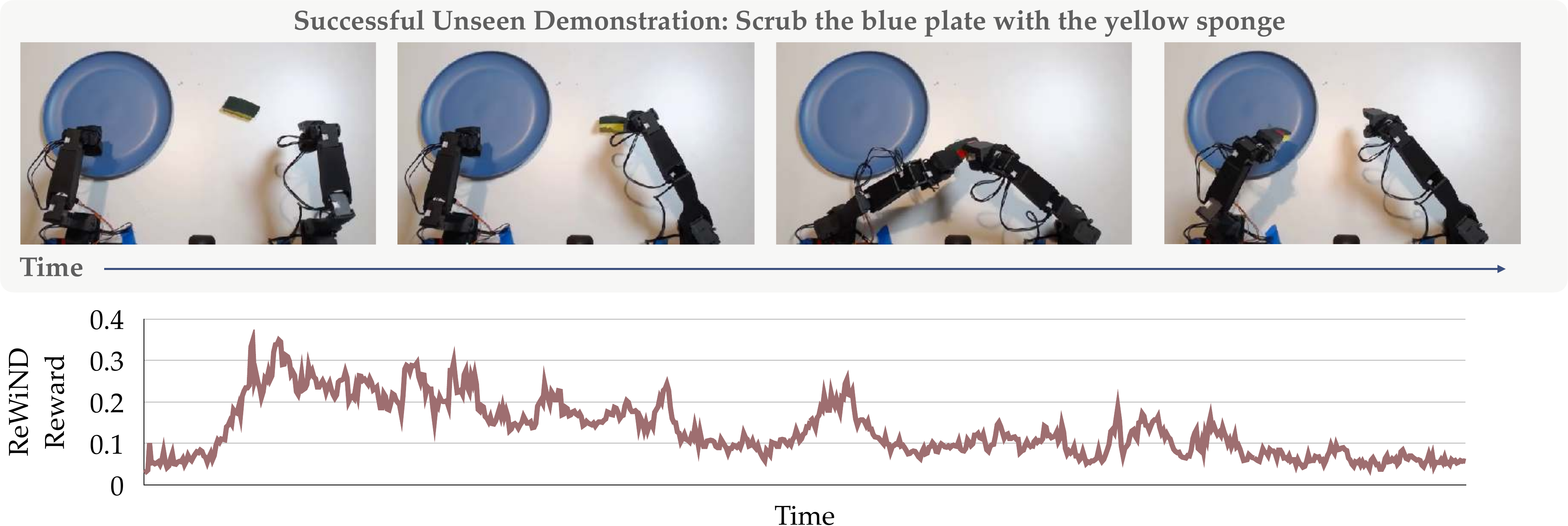}
    \caption{\textbf{\method\ Failure Example.} We collect a demonstration of the Koch arms picking up a sponge, handing it over, and scrubbing a plate. We find that there is poor reward alignment to this successful demonstration, likely due to the lack of bimanual data in Open-X, occlusion of the sponge, and poor camera viewpoints.}
    \label{fig:limitations_analysis}
\end{figure}
\section{Limitations}
\label{sec:experiments:limitations}
\newtext{Here we discuss limitations of \method\ and directions for future work.}
\paragraph{Initial Policy Performance.} 
Firstly, we use a relatively simple policy architecture that is trained from scratch for each of our domains.
We expect better performance by combining \method\ with stronger policy architectures capable of ingesting more data (e.g., pre-trained vision-language-action models~\citep{open_x_embodiment_rt_x_2023, kim2024openvla, brohan2023rt2visionlanguageactionmodelstransfer, geminiroboticsteam2025geminiroboticsbringingai, hamster2025}) that have better \emph{zero-shot performance} on new tasks to enable even more sample-efficient learning irrespective of reward function.

In fact, we confirmed experimentally that initial zero-shot performance was strongly indicative of how well the policy will learn a new task in our real-world experiments.
For instance, we did not find \method\ to help a policy that confidently performs the wrong task \newtext{due to a combination of the KL-constrained objective we use for offline to online learning in real world experiments (see \Cref{sec:appendix:real_robot:training} for details) and the fact that unlearning poor behaviors takes a significant amount of time.}
If the \method\ reward function could be combined with stronger policies that are easy to learn online in the loop, we hope it will enable learning of many more difficult new tasks.

However, the best way to fine-tune these models with online rewards remains an open challenge ~\citep{nakamoto2024steering, guo2025improvingvisionlanguageactionmodelonline}.
One bottleneck is simply that, even with low-rank adaptation techniques that prior work found to help train large policy architectures more efficiently~\citep{liu2024tail, guo2025improvingvisionlanguageactionmodelonline, kim2024openvla}, fine-tuning these models takes a lot of compute and real-world time that makes real-world online learning with reward difficult, albeit feasible in simulation~\citep{lu2025vlarlmasterfulgeneralrobotic}.
\cameraready{Recent work has also explored how to use RL to improve more expressive policy classes~\cite{wagenmaker2025steering, dong2025expo, chen2025conrft}.
We show preliminary experiments with Diffusion Steering RL~\cite{wagenmaker2025steering} with \method, along with discussions on potential drawbacks to such methods for learning \textit{new} tasks}.
We plan to investigate blending \method\ approaches with such policy architectures and efficient fine-tuning techniques, such as large model steering approaches~\citep{wagenmaker2025steering}, in the future.

\paragraph{Reward Objective.} One limitation of using video progress as a way to label training data, whether one uses relative targets~\citep{alakuijala2024videolanguagecritictransferablereward, yang2024rank} or fixed targets such as with \method\ in \Cref{eq:progress_loss}, is that we cannot explicitly incorporate collected \emph{failure trajectories} into \Cref{eq:progress_loss} without labels indicating what the final task progress for the trajectory should be.
\citet{alakuijala2024videolanguagecritictransferablereward} incorporates failure trajectories by using them as contrastive negatives while disabling their sequential ranking loss which relies on frame progress. 
Similarly, \method\ can na\"{i}vely be trained with collected failure trajectories by only using them as negatives with a progress target of 0 in \Cref{eq:progress_loss}.
However, future work can incorporate other objectives, use accurate VLMs, or assume human labelers to label these videos with approximate progress targets.

\paragraph{Reward Analysis.}
One of the limitations of \method\ lies in its inherent tradeoff using pre-trained vision and language embeddings.
We do not fine-tune these embeddings because our assumed demonstration dataset $\demos$ is very small, and in early experiments, we found that fine-tuning sometimes hurts generalization performance.
Not fine-tuning the frozen embeddings may result in the reward model generally underfitting certain tasks, particular to robotics, on which the pre-trained vision and language models were not trained.
In \Cref{fig:limitations_analysis}, we visualize an example of dish scrubbing which \method\ does not perform well on even though similar linguistic tasks exist in the Open-X dataset.
This poor result is likely due to Open-X not containing any bimanual data or partial occlusion due to the camera viewpoint.
Future work that pre-trains with even more robotics data or incorporates intermediate representations or objectives with large-scale pre-training on internet data (e.g., \citet{hamster2025, geminiroboticsteam2025geminiroboticsbringingai}) could allow fine-tuning the input embeddings to ensure they can better fit the $\demos$.

\paragraph{Resets.} \method, in its current form, requires a human operator to perform resets of the environment. 
This assumption prevents \method\ from being fully autonomous. However, recent reset-free RL works~\citep{yang2023robot, gupta2021reset, montgomery2017reset, ye2024reinforcement} 
demonstrate promising solutions to address the need for humans to supervise learning. 
Regardless, human resets remain a roadblock to autonomous learning that is difficult to address in the real world~\citep{mirchandani2024so}. 

\paragraph{Success Detection.} Another limitation comes from requiring success detection for the reward bonus and terminating policy rollouts upon success.
We add a success bonus (detailed in \Cref{sec:method:policy learning}) to account for potential noisy rewards and imperfect success detection by the reward model, given that a human is already monitoring to reset the environment, and \newtext{we subsequently} terminate the rollout upon success. 
Methods such as those introduced by \citet{ye2024reinforcement, zhou2024autonomous, yang2023robot},
which utilize VLMs as success detectors, can remove the need for human supervision during the online phase of \method\ when combined with reset-free RL.
In future work, we plan to investigate \method\ with reset-free approaches\cameraready{, guided by task performance estimates~\cite{anwar2025efficient} and} automatic success detection for truly autonomous learning.

\acknowledgments{
We thank Sid Kaushik for feedback on the final draft of the paper.
This work was supported in part by a grant from the Army Research Lab (ARL) Army AI Innovations Institute (A2I2), award number W911NF-23-2-0010. The claims and findings of this work do not necessarily represent the views of the ARL. This work was also supported in part by a grant from the DARPA Friction and Accountability in Conversational Transactions (FACT) program.
Additionally, this work was supported by an Institute of Information \& Communications Technology Planning \& Evaluation (IITP) grant (No.RS2019-II190075, Artificial Intelligence Graduate School Program, KAIST) and National Research Foundation of Korea (NRF) grant funded by the Korea government (MSIT) (NRF-2021H1D3A2A03103683, Brain Pool Research Program; RS-2024-00414822).
}

\clearpage
\bibliography{references}  %

\clearpage
\appendix
\begin{figure}[t]
   \centering
    \includegraphics[width=\linewidth]{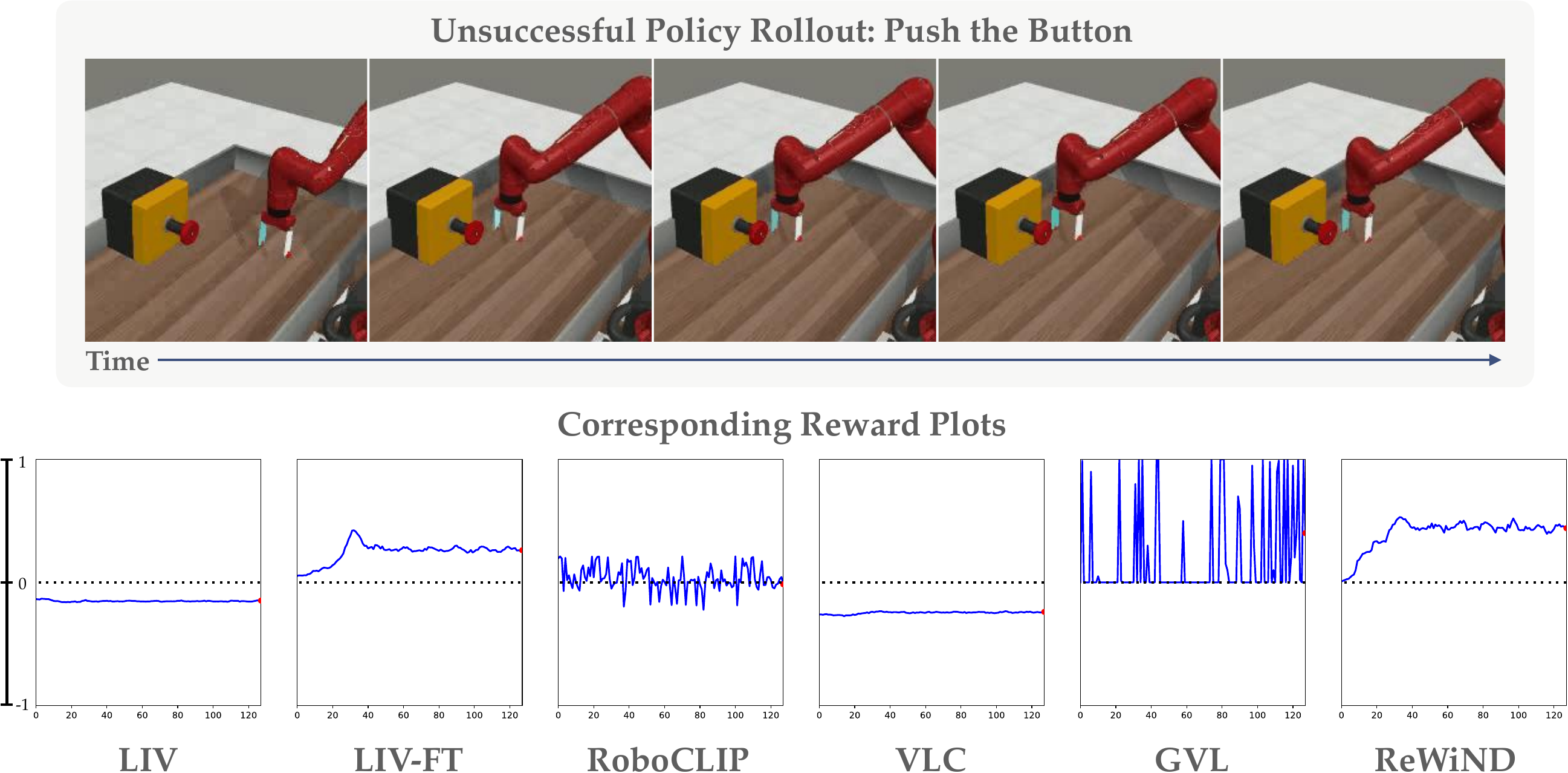}
    \caption{Unsuccessful policy rollout for the ``Push the Button'' task in Meta-World and its corresponding rewards below it. \method\ predicts calibrated rewards that reflect better partial progress when the policy gets stuck near the button.} %
    \label{fig:exps:policy_perf_analysis}
\end{figure}

\section{Implementation Details}
\label{sec:appendix:implementation}
This section introduces implementation details for \method\ in terms of the datasets, reward model, policy training, and online RL.

\subsection{\method\ Implementation}
\label{sec:appendix:implementation:rewind_iplmeentations}
\begin{algorithm}
\caption{\method\ Algorithm, \Cref{sec:method}.} \label{alg:full_approach_overview}
\begin{algorithmic}[1]

\Require Demo dataset $\demos$, Pre-trained LLM, Open-X subset $\openx$, Reward Model $\rewardgeneral$, Policy $\pi$. $\demos$ includes video trajectories \( o_{1:t} \) and language embedding $z$.

\State \textcolor{algblue}{\textit{/* Train the Reward Mode}l \Cref{sec:method:reward} */}
\State \textsc{RewardModelTraining}($\rewardgeneral$, $\demos$, $\openx$) 

\State \textcolor{algblue}{\textit{/* Policy Pretraining} \Cref{sec:method:policy learning} */}
\State \textsc{OfflinePolicyPretraining}($\rewardgeneral$, $\demos$, $\pi$) 
\State \textcolor{algblue}{\textit{/* Learn New Task Online} \Cref{sec:method:policy learning} */}
\State \textsc{OnlineRL}($z_\text{new}$, $\rewardgeneral$, $\pi$ )

\State 

\Procedure{RewardModelTraining}{$\rewardgeneral$, $\demos$, $\openx$} 
    \State Augment instruction labels with LLM
    \State Sample a video clip and annotation \( o_{t_1:t_2}, z \) from $\demos$ or $\openx$.
        
    \State Choose to keep the original video or perform \textsc{RewindAugmentation}.
            \If {perform \textsc{RewindAugmentation}}
                \State $o^\text{rewound} \gets $ \textsc{RewindAugmentation}$(o_{t_1:t_2})$
                \State Optimize $\rewardgeneral$ with $\mathcal{L}_\text{rewind}(o^\text{rewound}, z)$ \Comment{\Cref{eq:rewind}}
            \Else
                \State Sample a different video clip $o_{t_1':t_2'}^\text{other}$
                \State Optimize $\rewardgeneral$ with $\mathcal{L}_\text{progress}(o_{t_1:t_2}, z, o^\text{other}_{t'_1, t'_2})$ \Comment{\Cref{eq:progress_loss}}
            \EndIf  
\EndProcedure

\State 
\Procedure{OfflinePolicyPretraining}{$\rewardgeneral$, $\demos$, $\pi$} 
    \State Relabel $\demos$ with $\offlinerewards$ coming from $\rewardgeneral$. \Comment{\Cref{eq:offline_labeled_rewards}} %
    \State Train $\pi$ with offline RL on relabeled $\demos$.
\EndProcedure
\State 
\Procedure{OnlineRL}{$\rewardgeneral$, $\pi$}
    \State For every rollout label the trajectories with $\onlinerewards$ from $\rewardgeneral$. \Comment{\Cref{eq:online_rewards}}
    \State Optimize $\pi$ with online RL Algorithm
\EndProcedure
\State 
\Procedure{RewindAugmentation}{$o_{t_1:t_2}$} \Comment{\Cref{sec:method:reward:augmentation}}
    \State Sample random split point $i$ between $t_1$ and $t_2$.
    \State Sample \# frames to rewind for, $k$
    \State Reverse $o_{i-k:i}$ and concat with $o_{t_1: i}$
    \State Return $[o_{t_1:i-1}, o_{i:i-k}]$
\EndProcedure

\end{algorithmic}
\end{algorithm}

Full pseudocode for \method\ is listed in \Cref{alg:full_approach_overview}.
Individual implementation details follow.
\subsubsection{Open-X Dataset}
\label{sec:appendix:dataset}
Below we list details of the OXE subset, $\openx$, used for training the reward model $\rewardgeneral$ (mentioned in \Cref{sec:method:reward}).

We select a subset of datasets from the Open-X Dataset~\citep{open_x_embodiment_rt_x_2023}.
The subset includes Bridge-V2~\citep{walke2023bridgedata}, BC-Z~\citep{jang2021bcz}, Fractal~\cite{brohan2022rt1}, CLVR Jaco Play~\citep{dass2023jacoplay}, Berkeley Autolab UR5~\citep{BerkeleyUR5Website}, Berkeley Fanuc Manipulation~\citep{fanuc_manipulation2023}, CMU Stretch~\citep{bahl2023affordances, mendonca2023structured}, Stanford Hydra~\citep{belkhale2023hydra}, UCSD Kitchen~\citep{ucsd_kitchens}, Austin BUDS~\citep{austinbuds}, Austin Sirius~\citep{liu2022robot}, and DROID~\citep{khazatsky2024droid}.
These datasets were selected for their high-quality, task-oriented manipulation trajectories (i.e., no play data or extremely high-level annotations).
These datasets provide around 350k trajectories and 58k total unique task annotations. To ensure meaningful trajectories for training the \method\ reward model, we postprocess the data to remove trajectories with less than 5 timesteps. 
We subsample the videos in the datasets to 16 frames for reward model training, as we did not see a noticeable benefit from training it with longer videos.

\subsubsection{Reward Function}
\label{sec:appendix:implementation:rewind_implementations:reward}
\begin{figure}[t]
    \centering
    \includegraphics[width=\linewidth]{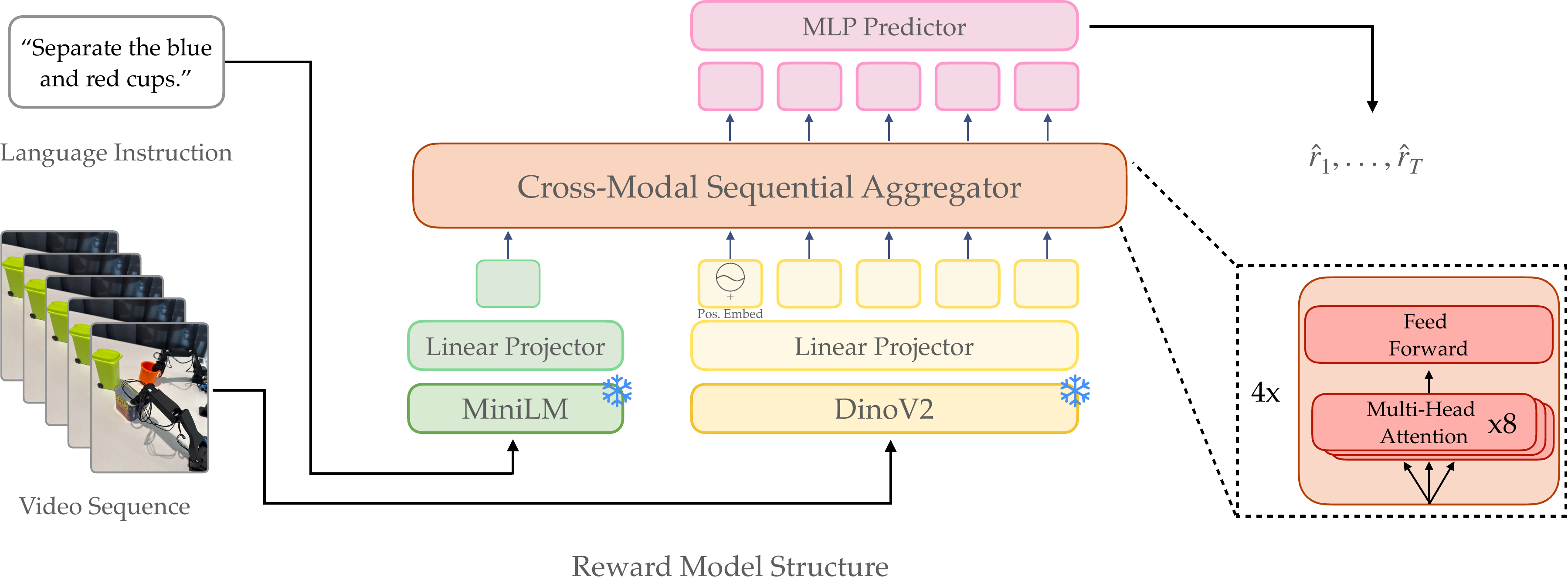}
    \caption{\textbf{\method's Reward Model Architecture}. It's composed of frozen language and image input embeddings projected to a shared hidden dimension of 512. These embeddings are treated as input tokens to the \outputhead\ transformer composed of 4 causally masked transformer layers composed of 8 multi-head attention blocks each. Per-timestep embeddings for each input observation are fed into an MLP to predict rewards for each timestep.}
    \label{fig:reward_model_architecture}
\end{figure}
We picture the overall architecture of the reward function in \Cref{fig:reward_model_architecture}.
We encode input images with the pre-trained \textsc{DINO-v2} base model (86M params) with 768 embedding size.
Similarly, we encode language with the pre-trained \textsc{all-minilm-l12-v2} model with a 384 embedding size. We project image and language embeddings to 512 dimensions with a single linear layer. 
We treat the language embedding as a single input token and we evenly downsample DINO-v2 image embeddings for every observation sequence to 16 frames.

The \outputhead\ takes these tokens as input and produces a per-image embedding used by an MLP to produce per-timestep rewards. 
The \outputhead\ is a causally masked transformer (\texttt{PyTorch nn.TransformerEncoder}) composed of 4 layers, each with 8 heads with a combined hidden dimension of 2048.
We add a learnable positional embedding to only the first frame of the video sequence embedding. 
In the ReWind reward function training phase, we trained 2k steps for Meta-World and 10k steps for Real-World robot experiments, with a batch size of 1024. Each batch includes 80\% data from $\openx$ and 20\% target environment data from $\demos$. 
Each video in the batch has an 80\% probability of having video rewind augmentation, and independently, a 20\% percent probability of having a mismatched video-language pairing with 0 progress target (see \Cref{sec:method:reward}). 
\cameraready{For the rewind augmentation, the split point is uniformly sampled from the video clip, and the rewound sequence is obtained by taking all frames from the split point to the end of the clip.}
In order to better simulate policy execution videos that look \emph{almost successful}, 10\% of the rewound videos will only have their last 3 frames rewound. 
No extensive tuning was performed on these per-sample rewind and mismatch probabilities; they were heuristically chosen during initial small-scale experimentation and then fixed for all experiments.

\subsubsection{Policy Training}
\label{subsec:appendix:implementation:policy_training}
Specific architectural and training details are discussed per-environment in the corresponding sections \Cref{sec:appendix:metaworld:training} and \Cref{sec:appendix:real_robot:training}. 
Below we talk about high-level algorithmic details for policy training along with shared implementation details across environments.
\paragraph{Policy Input.}
Similar to the reward model, we condition the policy on frozen pre-trained image and language embeddings: DINO-v2-base image embeddings (86M params, 768 embedding size)~\citep{oquab2024dinov2learningrobustvisual} along with \textsc{all-minilm-l12-v2} language embeddings of size 384 from the Sentence Transformers Python package~\citep{reimers-2019-sentence-bert}.
\newtext{We also include proprioceptive information in both of our envrionments.}

\paragraph{Offline RL.}
We use Implicit Q Learning (IQL)~\citep{kostrikov2022offline} as prior work found it performant and easy to tune for robot manipulation with action-chunked policies~\citep{zhang2023bootstrap, zhang2024sprint, wu2024vformer}. 
IQL trains on in-distribution $(s, a, s', r, a')$ tuples from the dataset, avoiding using \newtext{next actions} $a'$ sampled from a policy, to ensure the critic functions accurately reflect returns restricted to dataset actions. The value function is optimized with expectile regression, controlled by a hyperparameter $\tau$: $\tau = 0.5$ recovers mean squared error, while $\tau \to 1$ yields a more optimistic estimate, helping the value function ``stitch'' together distant rewards in sparse settings. The policy is trained via advantage-weighted regression~\citep{awr}, maximizing
$$
e^{\beta (Q(s, a) - V(s))} \log \pi(a|s),
$$
where $\beta$ is a temperature hyperparameter controlling how ``spiky'' the policy loss is. 
To prevent numerical instability, the exponential term is capped at a maximum value in practice (for us, this is 100).

\paragraph{Online RL.}
We use a custom soft-actor critic (SAC)~\citep{haarnoja2017soft} implementation initialized with the pre-trained policy from offline RL along with the Q and target Q functions.
We follow best practices from recent offline-online RL fine-tuning work~\citep{zhou2024efficientonlinereinforcementlearning, ball2023rlpd}, namely:
\begin{itemize}
    \item 5-10 critics instead of 2, with random sampling of critics
    \item LayerNorm in the critic and possibly LayerNorm in the policy
    \item A higher update-to-data ratio in the critics
    \item ``Warm-starting'' online RL by running with the frozen pre-trained policy for the first few thousand environment steps~\citep{zhou2024efficientonlinereinforcementlearning, uchendu2023jumpstartreinforcementlearning}
    \item Possibly sampling offline pre-training data at a 50\% ratio during online RL
    \item Removing the SAC entropy term from the target critic
\end{itemize}

We found that by default, efficient offline-online learning algorithms did not work very well ``out of the box'' for learning \emph{new} tasks on our real robot. 
This is perhaps because they focus specifically on offline-online fine-tuning on the same task while we are trying to learn new tasks, or perhaps due to additional challenges of real-robot RL.
Therefore, we make some per-environment design decisions for online RL detailed in the respective environment training sections.

\section{MetaWorld Experiments}
\label{sec:appendix:metaworld}

\subsection{Simulation Setup}
\label{sec:appendix:metaworld:setup}

\paragraph{Training/Eval Task Selection.} 
We manually select 20 training tasks from MT50 benchmark in the MetaWorld environment. These tasks are used for both reward model training and policy pre-training. 
The training tasks include: \texttt{Button-Press, Button-Press-Topdown-Wall, Coffee-Pull, Dial-Turn, Door-Open, Door-Unlock, Drawer-Close, Faucet-Open, Handle-Press, Handle-Pull-Side, Peg-Insert-Side, Pick-Place, Plate-Slide, Plate-Slide-Back-Side, Push, Reach, Stick-Push, Stick-Pull, Window-Open, Hand-Insert}.

We also choose another 17 tasks from the MT50 benchmark for reward model evaluation and 8 of tasks are selected for downstream policy finetuning.\footnote{These 17 tasks were chosen for sharing at least some characteristic with a training task.} The evaluation tasks include \texttt{Window-Close, Sweep-Into, Soccer, Reach-Wall, Push-Back, Plate-Slide-Side, Plate-Slide-Back, Pick-Place-Wall, Handle-Pull, Handle-Press-Side, Faucet-Close, Door-Lock, Door-Close, Coffee-Push, Coffee-Button, Button-Press-Wall, Button-Press-Topdown}. 
\cameraready{We select these training tasks and target evaluation tasks to test directional and spatial (e.g., door open → close, button press → + wall) generalization scenarios where the policy is unlikely to be able to solve the task, due to requiring significantly different action sequences from the most visually similar training task, while a good reward function should be able to guide the policy to solve it.}
The 8 tasks used for downstream policy training are \texttt{Window-Close, Reach-Wall, Faucet-Close, Coffee-Button, Button-Press-Wall, Door-Lock, Handle-Press-Side, Sweep-into.}

\paragraph{Environment Details.}
\begin{wrapfigure}{R}{0.3\textwidth}
\centering
\includegraphics[width=\linewidth]{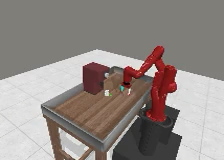}
\caption{Example camera viewpoint in MetaWorld.}
\label{fig:metaworld view}
\end{wrapfigure}
We use MetaWorld~\citep{yu2021metaworldbenchmarkevaluationmultitask} with the default 3rd-person camera viewpoint, pictured in \Cref{fig:metaworld view}, and also 4-dimension proprioception input ($x, y, z, \text{gripper}$).
The policy action space is the default one from MetaWorld represented as a 4-dimensional relative action space for ($\Delta x, \Delta y, \Delta z, \text{gripper}$). 
Unlike the MetaWorld environment setups in prior reward learning papers, we do not include goal/ground truth state information. 
We also terminate the environment on success.
Both of these choices were made to mimic a real-world robot learning setup.
The time horizon of each episode is limited to 128 steps.
\newtext{The success bonus for online and offline RL used in \Cref{eq:offline_labeled_rewards} and \Cref{eq:online_rewards} is 200 for \method\ and all baselines.}

\subsection{Training Details}
\label{sec:appendix:metaworld:training}
For $\demos$, we select 20 tasks from the MT-50 benchmark. Each task consists of one human-labeled annotation, four augmented annotations (\Cref{sec:method:reward:augmentation}), and five optimal \newtext{demonstrations} produced by the MetaWorld built-in planner.
We render images at the default resolution of 640x480, centercrop to 224x224 and embed the image with DINOv2 encoder. 

We pre-train the policy with IQL \citep{kostrikov2022offline} for 100K steps with learning rate 0.001, discount $\gamma=0.99$. We use a three-layer MLP of size $[768, 512, 256]$ for both the policy and value function network. 
The general training procedure is described in \Cref{subsec:appendix:implementation:policy_training}

For the various hyperparameters for online policy learning we used in MetaWorld as described in Section~\ref{subsec:appendix:implementation:policy_training}.
We use 10 critics and sample 2 of them during training, LayerNorm in both the critic and policy, and an update-to-data ratio of 4 for the critics. 
We are not sampling from offline pre-training data during online \newtext{training} nor are we training the target critic with the entropy term\newtext{, so the implementation is identical to Warm-Start RL~\citep{zhou2024efficientonlinereinforcementlearning}}.
We warm-start online RL for 4000 steps.

\section{Real Robot Experiments}
\label{sec:appendix:real_robot}

\subsection{Robot Experiment Setup}
\label{sec:appendix:real_robot:setup}
\begin{figure}
    \centering
    \includegraphics[width=\linewidth]{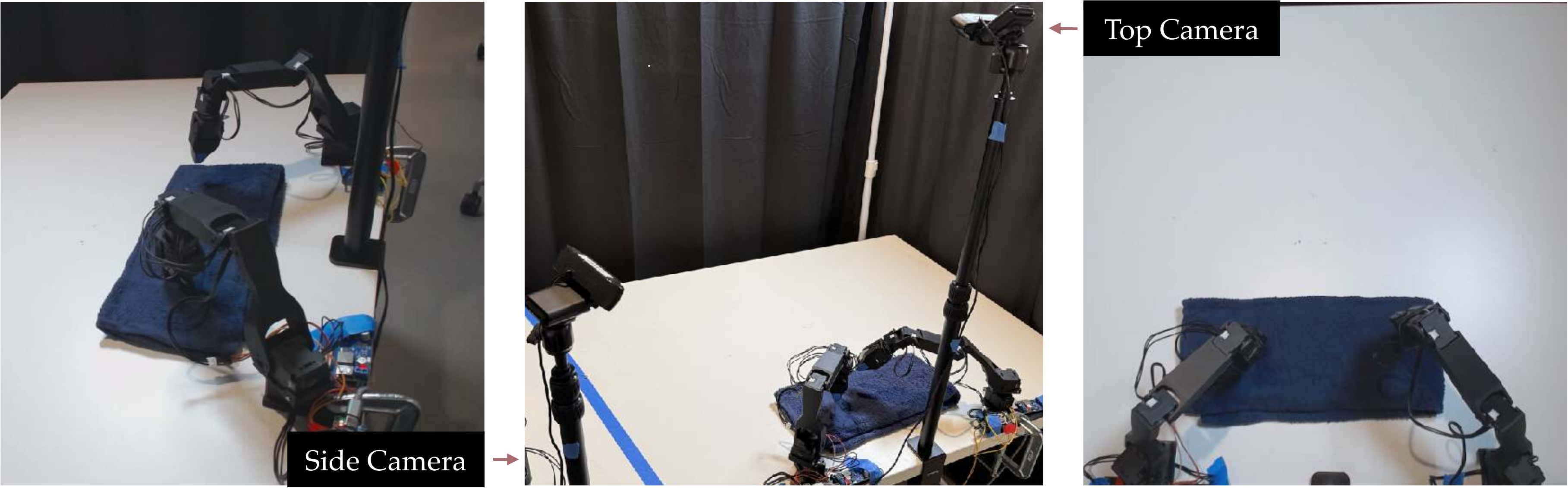}
    \caption{\textbf{Real World Bimanual Robot Setup.}
    Our real-world setup consists of a top-down and side camera mounted to a table where two Koch v1.1 low-cost arms are mounted. This setup allows us to perform bimanual tasks and easily collect data with another pair of low-cost ``leader'' arms mounted to the same table. 
    }
    \label{fig:koch_view}
\end{figure}
We use the Koch1.1 bimanual arm setup for data collection and learning~\citep{cadene2024lerobot}.\footnote{\url{https://github.com/jess-moss/koch-v1-1}}
Altogether, four total arms (2 for data collection) cost $\sim$\$1000, letting us demonstrate \method\ enables real-world online RL of new tasks even with very low-cost hardware and noisy control.
The observations consist of RGB images from a Logitech C930e top camera and side camera (pictured in \Cref{fig:koch_view}). 
We control the robot with absolute joint position control at a frequency of 30Hz.
We collect a small dataset of 10 demonstrations over 20 tasks, and then use 5 demos per-task for the reward function.
We found the offline-trained policy to be the primary bottleneck to optimizing rewards in unseen tasks, so we used 10 demos per-task for offline policy training.
\newtext{We have an episode timeout of 250 steps and provide a success bonus of 125 upon success (from \Cref{eq:offline_labeled_rewards} and \Cref{eq:online_rewards}).}
\newtext{Proprioceptive information in this environment includes 12 robot joint states, 6 for each arm.
These represent the rotation of each joint and gripper.}

\subsection{Real Robot Training Details}
\label{sec:appendix:real_robot:training}
We use a small, instruction-tuned, open-source LLM, \texttt{Mistral-7B-Instruct-v0.3}~\citep{jiang2023mistral7b}, to generate 9 additional instructions for each task for instruction augmentation.

For the small dataset in real robot experiments, we manually choose 15 tasks in the Koch tabletop setting, and each task includes 5 trajectories and 10 annotations. The evaluation set is 5 other random tasks, which are irrelevant with the tasks in the small dataset. 
We use this evaluation set for offline metrics and validating various design choices.

Unlike the MetaWorld experiments that use an MLP-based policy, we use an action-chunked policy with temporal ensembling for the real robot. 
We found chunking to lead to more stable bimanual manipulation on the Koch arms.
We implement the action chunking with a Transformer policy that predicts 60 actions at each timestep corresponding to 2 seconds of actions.
We also implement a Transformer-based critic.
During rollouts, we then use temporal ensembling~\cite{zhao2023learning}.
Here, the current action is ensembled with the last 60 timesteps' predictions according to an exponential weighting scheme $w_i=\exp(-m*i)$, where we use $m=0.01$ or $m=0.1$ depending on the task. \newtext{We found $m=0.1$ to work well for tasks requiring grasping solid objects as it weights recent actions more heavily, necessary for ensuring the policy actually commits to the grasp, and $m=0.01$ to work well for non-grasping tasks as it results in a smoother policy.}

\newtext{The policy is a Transformer decoder with 1 layer and 8 heads with 1.5M params. The critic is a Transformer encoder with 8 heads and 1 layer.}
We train each policy for 20k steps offline on our offline dataset using IQL with AWR for policy extraction.
We train using a batch size of 256, use 5 critics, and subsample 2 critics at each training step.
\newtext{We use LayerNorm only in the critics as we found that LayerNorm in the action-chunked policy could potentially hurt RL performance.}
We also warm-start online RL for 3000 steps.
\newtext{We do not sample actions during policy rollouts as we found action sampling to conflict with temporal ensembling.}

Then, we train the policy online as described in Section~\ref{sec:method:policy learning}.
We train online for 50k \newtext{environment} steps, which takes approximately 1 hour \newtext{as there is minimal waiting time for policy training due to a threaded implementation that trains the policy while the last iteration's policy checkpoint is used for rollouts.}
This parallelization nearly doubles the rate at which we are able to collect policy rollouts.
Specifically, during online training, we collect a single rollout \newtext{corresponding to 250 environment steps while simultaneously training the policy for 75 gradient steps.} 
\newtext{We keep a relatively low policy to environment update ratio in order to ensure that we do not have to wait for offline training to finish in order to start the next online rollout.}
At each gradient step, we sample our buffer such that 50\% is the offline training data, 25\% is online failure trajectories, and 25\% is online successful trajectories. 
This sampling approach helps upsample successful online trajectories.
For every actor gradient step, we do \newtext{5} critic update steps to more quickly train the critic online.

During real-world policy rollouts, it is important for the robot to take safe actions that will not crash into other objects or the table.
However, we found that when regularizing the policy's KL divergence against a max-entropy prior as is the case in the entropy maximization objective in standard SAC~\citep{haarnoja2017soft}, the growing entropy term would cause the policy to produce largely random actions. 
Therefore, we regularize against the pretrained policy's distribution to encourage reasonable behaviors throughout the process of learning, \newtext{similar to the SAC update rules from \citet{pertsch2020spirl}}.
Thus the $\pi$ and $Q$ updates follow:
\begin{align}
\pi &\leftarrow \max_\pi \mathbb{E}_\pi \Big[Q(o, a, z) - \alpha \underbrace{\text{KL}(\pi(\cdot | o, z) \mid \mid \pi_\text{pretrained} (\cdot | o, z)}_{\text{pre-trained policy guidance}} \Big] \\
Q &\leftarrow \min_{Q} Q(o, a, z) = r + \gamma \Big[ Q(o', a, z) - \alpha \underbrace{\text{KL}\left(\pi(\cdot | o, z) \mid \mid \pi_\text{pretrained} (\cdot | o, z) \right)}_{\newtext{\text{pre-trained policy guidance}}}\Big]
\end{align}
We set $\alpha$ in both equations to a fixed value of 10.0 on \newtext{tasks where grasping solid objects is not required}.
\newtext{For others, we set it to 20.0 to ensure the policy doesn't degenerate from its grasping action early in training.}
We found that lower KL penalties could result in the policy falling into locally optimal but globally suboptimal behaviors, such as moving a cup with the arm instead of actually picking it up.
\cameraready{Work released after our paper further explores RL in chunked action spaces with similar KL-regularized behavior constraints, and find that such constraints and chunking helps improve exploration during RL~\cite{li2025reinforcement}}

\subsection{Real Robot Tasks}

We collected 10 demos per-task over 20 tasks on the Koch arms. We train the reward function on 5 demos per-task and the policy on 10 demos per-task. 
\cameraready{We selected these training and evaluation tasks specifically to evaluate spatial, visual, and language generalization.}
We list these training tasks below.

\texttt{Move the orange cup from the left to the right}, 
\texttt{Move the orange cup from the right to the left}, 
\texttt{Put the orange cup on the red plate}, 
\texttt{Put the red cup on the red plate}, 
\texttt{Separate the blue and red cups}, 
\texttt{Fold the blue towel}, 
\texttt{Open the green trash bin}, 
\texttt{Open the blue trash bin}, 
\texttt{Throw the banana away in the green trash bin}, 
\texttt{Throw the banana away in the blue trash bin}, 
\texttt{Put the red marker in the red trash can}, 
\texttt{Put the pink marker in the green trash can}, 
\texttt{Put the blue tape in the box on the left}, 
\texttt{Put the banana in the box}, 
\texttt{Put the orange cup in the box},
\texttt{Put the blue cup on the red plate}, 
\texttt{Separate the orange and blue cups}, 
\texttt{Open the red trash bin}, 
\texttt{Throw the banana away in the red trash bin}, 
\texttt{Put the red tape in the box on the right}.

\begin{figure}[t]
    \centering
    \includegraphics[width=\textwidth]{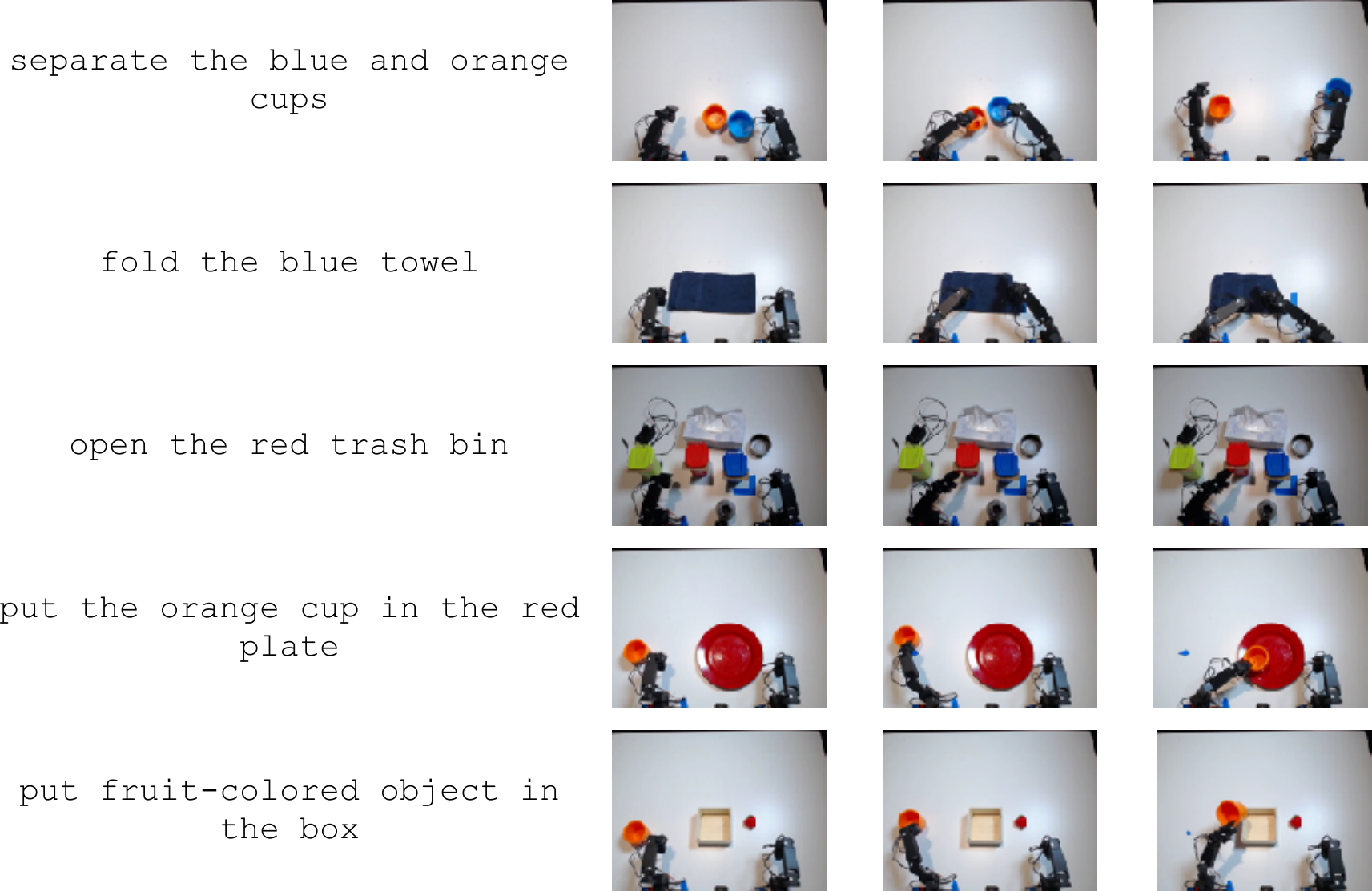}
    \caption{We present rollouts for the 5 tasks we use for online RL. The first two tasks are in-distribution to the policy, while the latter 3 tasks are out-of-distribution with respect to visual, spatial, or semantic generalization.
    }
    \label{fig:exps:rollouts}
\end{figure}

In addition, we present rollouts of the five online tasks in Figure~\ref{fig:exps:rollouts}. 
We also provide additional descriptions of these tasks below:

\begin{itemize}
    \item \texttt{Separate the blue and orange cups}: the robot must separate the two cups in the middle
    \item \texttt{Fold the blue towel}: the robot must fold the towel in half.
    \item \texttt{Open the red trash bin}: the robot is surrounded by clutter compared to the training data above and must open the trash bin
    \item \texttt{Put the orange cup in the red plate}: the robot picks an orange cup and must place it on a plate that is further away from the training data distribution
    \item \texttt{Put fruit-colored object in the box}: we refer to a ``fruit-colored" object to test the robot's ability to handle semantic generalization.
\end{itemize}

\section{Additional Results}
\label{sec:appendix:additional_results}

\subsection{Additional MetaWorld Reward Analysis}
\label{sec:appendix:additional_results:reward_analysis}
\begin{figure}[t]
    \centering
    \includegraphics[width=\linewidth]{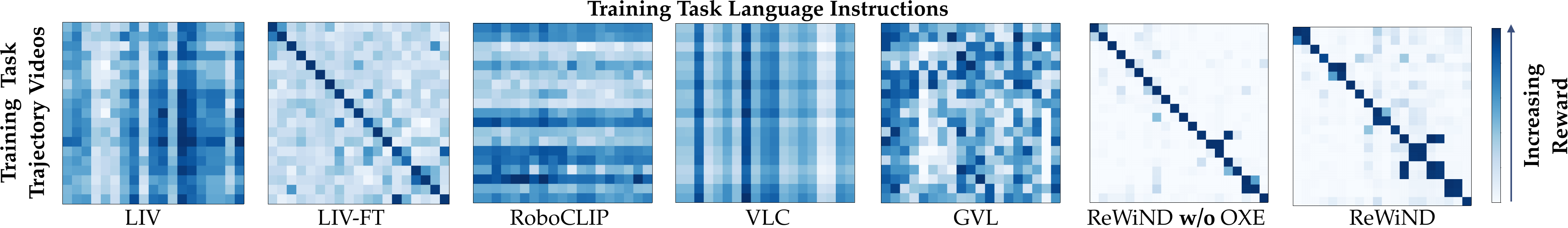}
    \vspace{-0.4cm}
        \caption{\textbf{MetaWorld Reward Confusion Matrix on 20 Training Tasks.} 
        For each training task, we compute rewards for all combinations of demonstration videos and language descriptions. 
        \method\ produces the most \textcolor{blue}{diagonal-heavy} confusion matrix, indicating strong alignment between unseen demos and instructions.
        }
    \label{fig:train_metaworld_confusion_matrices}
\end{figure}

In \Cref{fig:train_metaworld_confusion_matrices} we plot the confusion matrices of different reward models on training tasks in addition to the evaluation task plots of MetaWorld in \Cref{fig:exps:confusion_matrices}.  LIV, RoboCLIP and GVL are not pretrained or fine-tuned on the etraining tasks while VLC, LIV-FT and \method\ are. We can see both \method\ w/ OXE data $\openx$ and \method\ w/o OXE data $\openx$ are the best, having the clearest disparity between the diagonal and off-diagonal elements. LIV-FT also works well with a diagonal-heavy matrix. However, its disparity is not as clear as ReWiND.

\subsection{MetaWorld Sample Efficiency Results}
\label{sec:appendix:addtional_results:metaworld_sample_efficiency}

\begin{figure}[t]
  \centering

  \begin{subfigure}[b]{0.32\textwidth}
    \includegraphics[width=\linewidth]{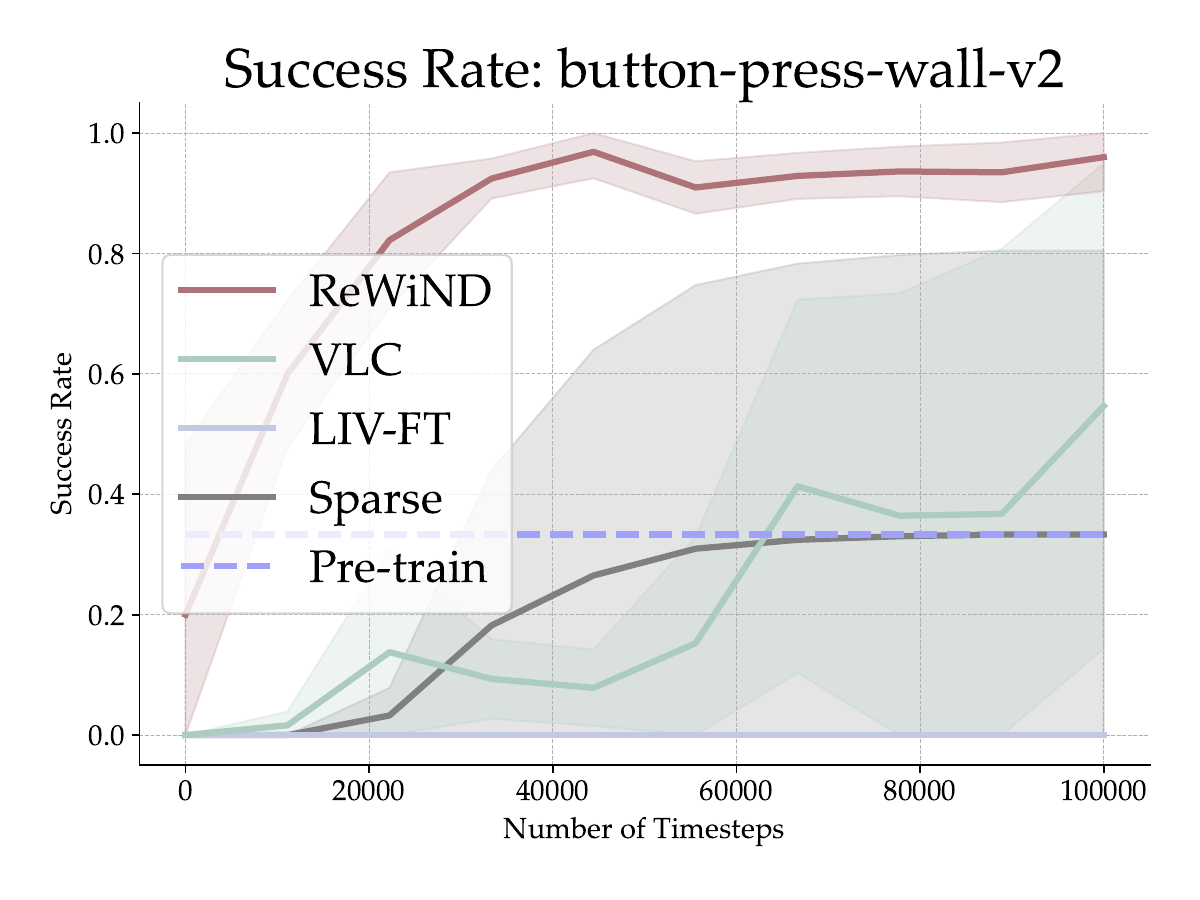}
    \caption{Button Press (Wall)}
  \end{subfigure}
  \hfill
  \begin{subfigure}[b]{0.32\textwidth}
    \includegraphics[width=\linewidth]{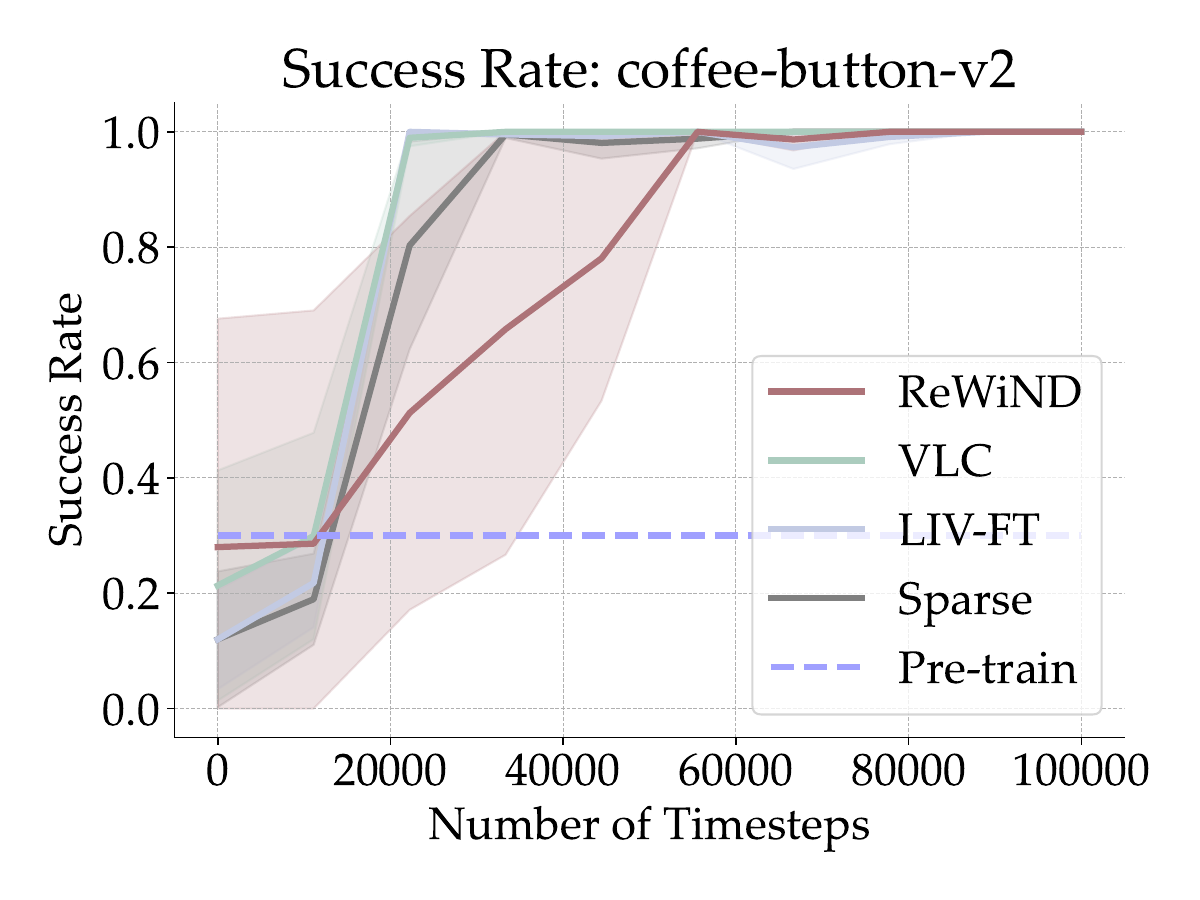}
    \caption{Coffee Button}
  \end{subfigure}
  \hfill
  \begin{subfigure}[b]{0.32\textwidth}
    \includegraphics[width=\linewidth]{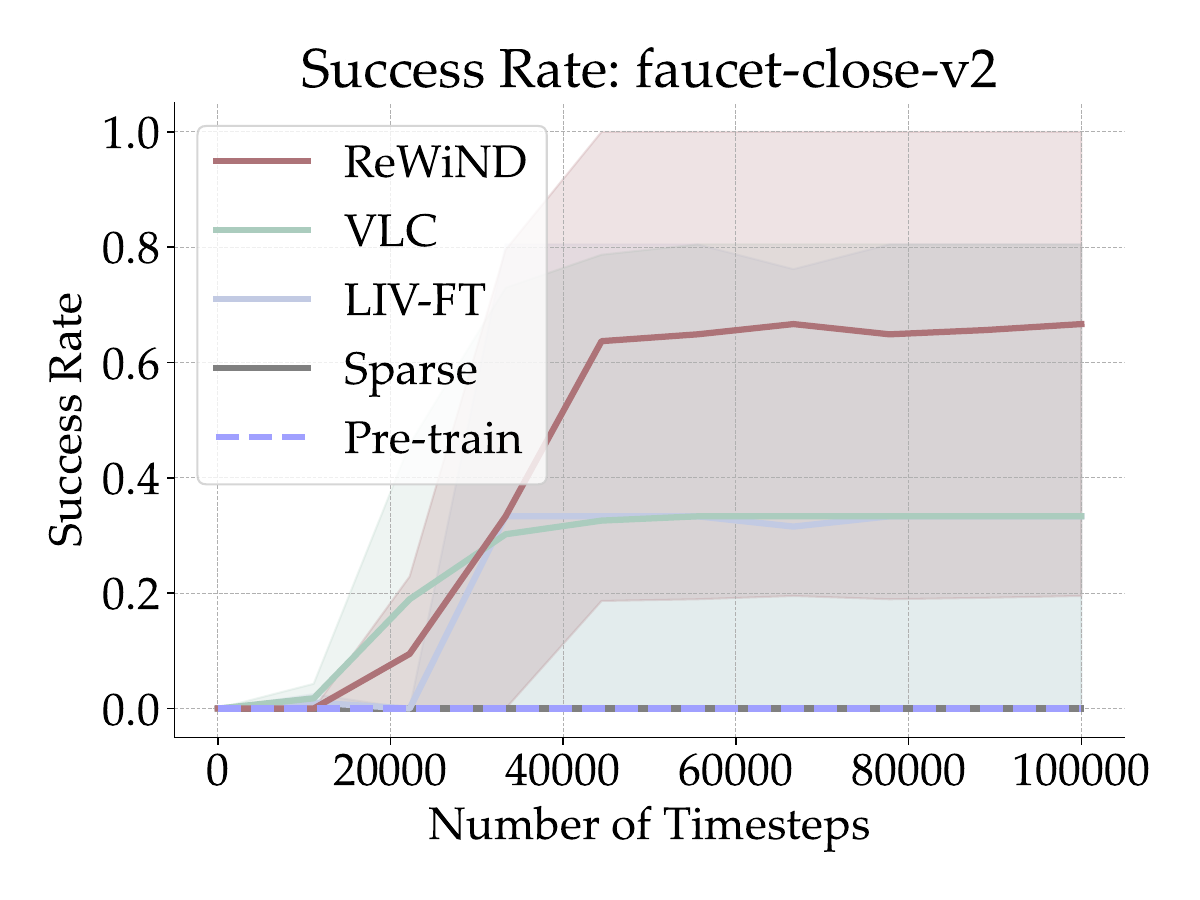}
    \caption{Faucet Close}
  \end{subfigure}

  \vspace{1em}

  \begin{subfigure}[b]{0.32\textwidth}
    \includegraphics[width=\linewidth]{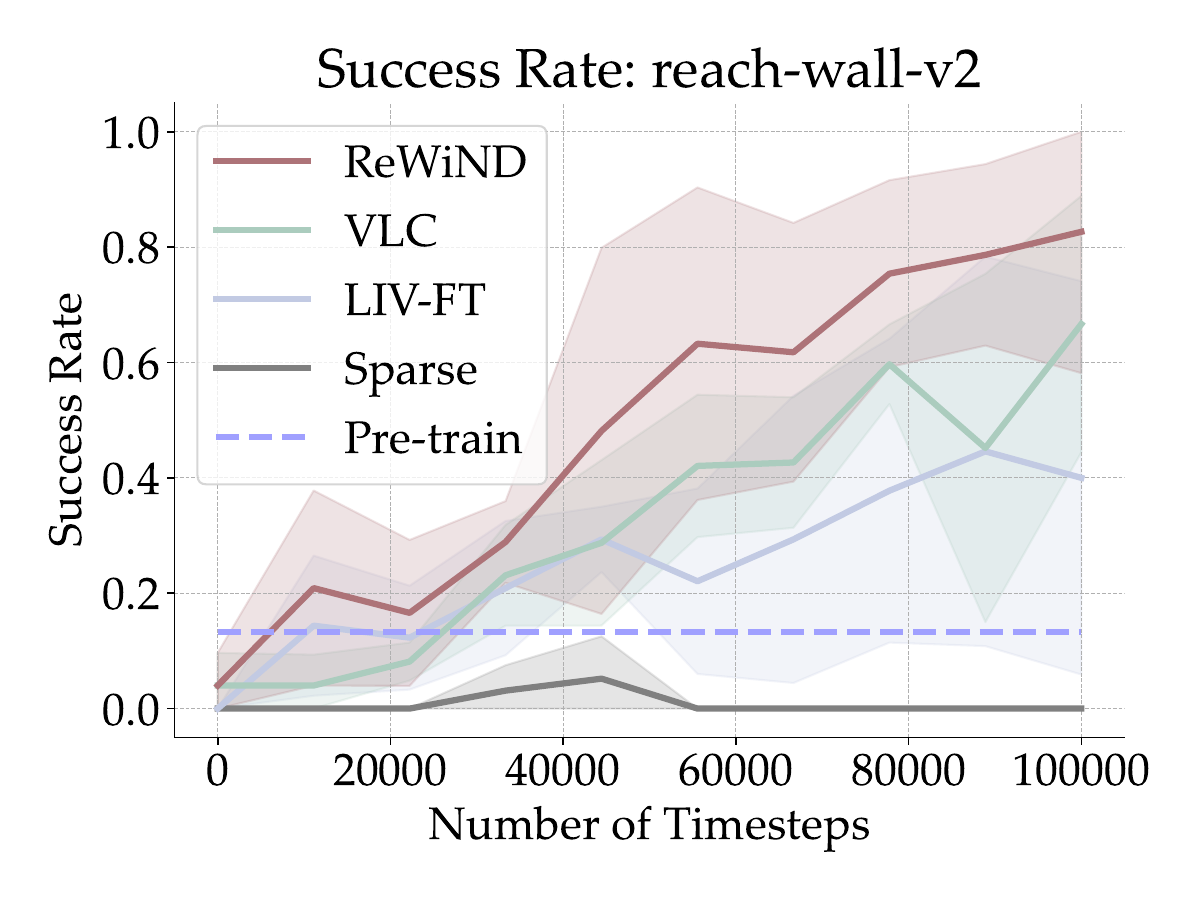}
    \caption{Reach (Wall)}
  \end{subfigure}
  \hfill
  \begin{subfigure}[b]{0.32\textwidth}
    \includegraphics[width=\linewidth]{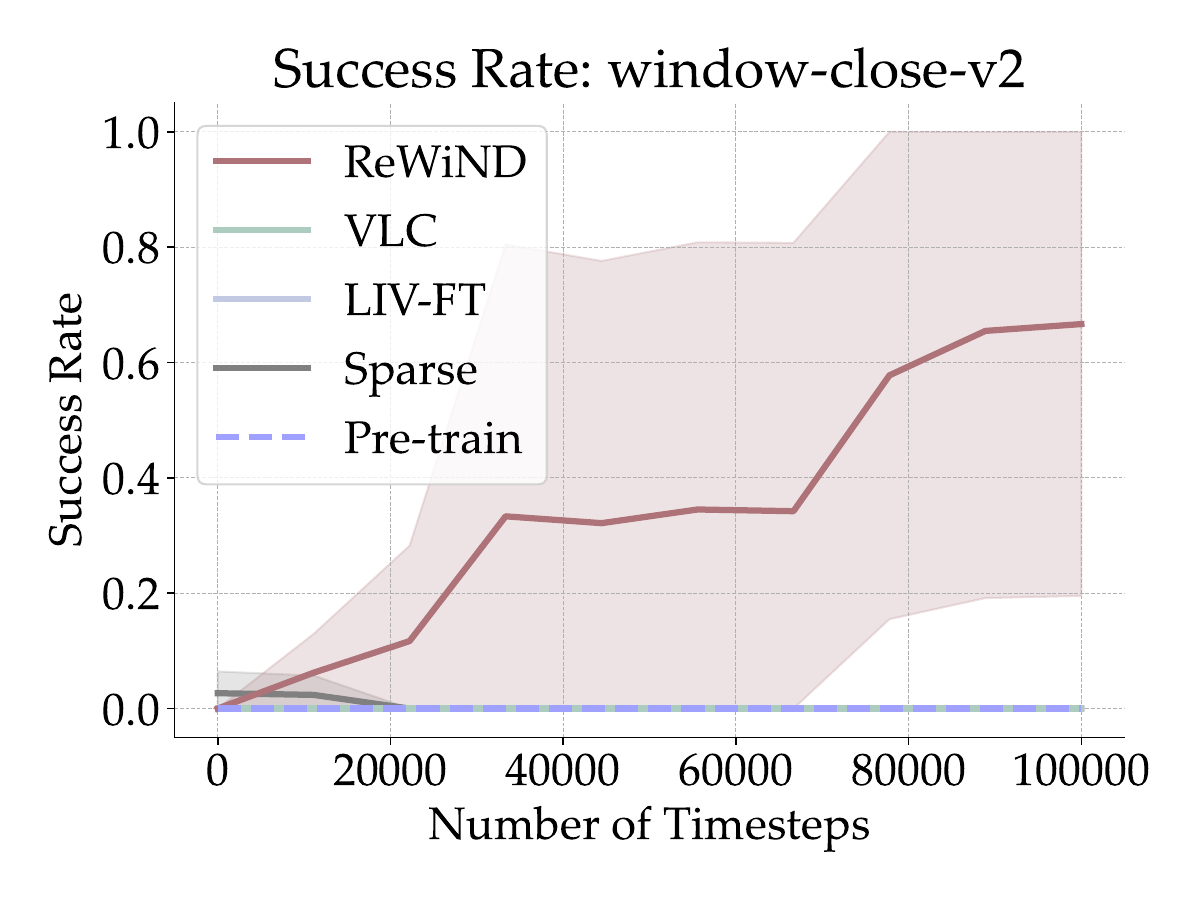}
    \caption{Window Close}
  \end{subfigure}
  \hfill
  \begin{subfigure}[b]{0.32\textwidth}
    \includegraphics[width=\linewidth]{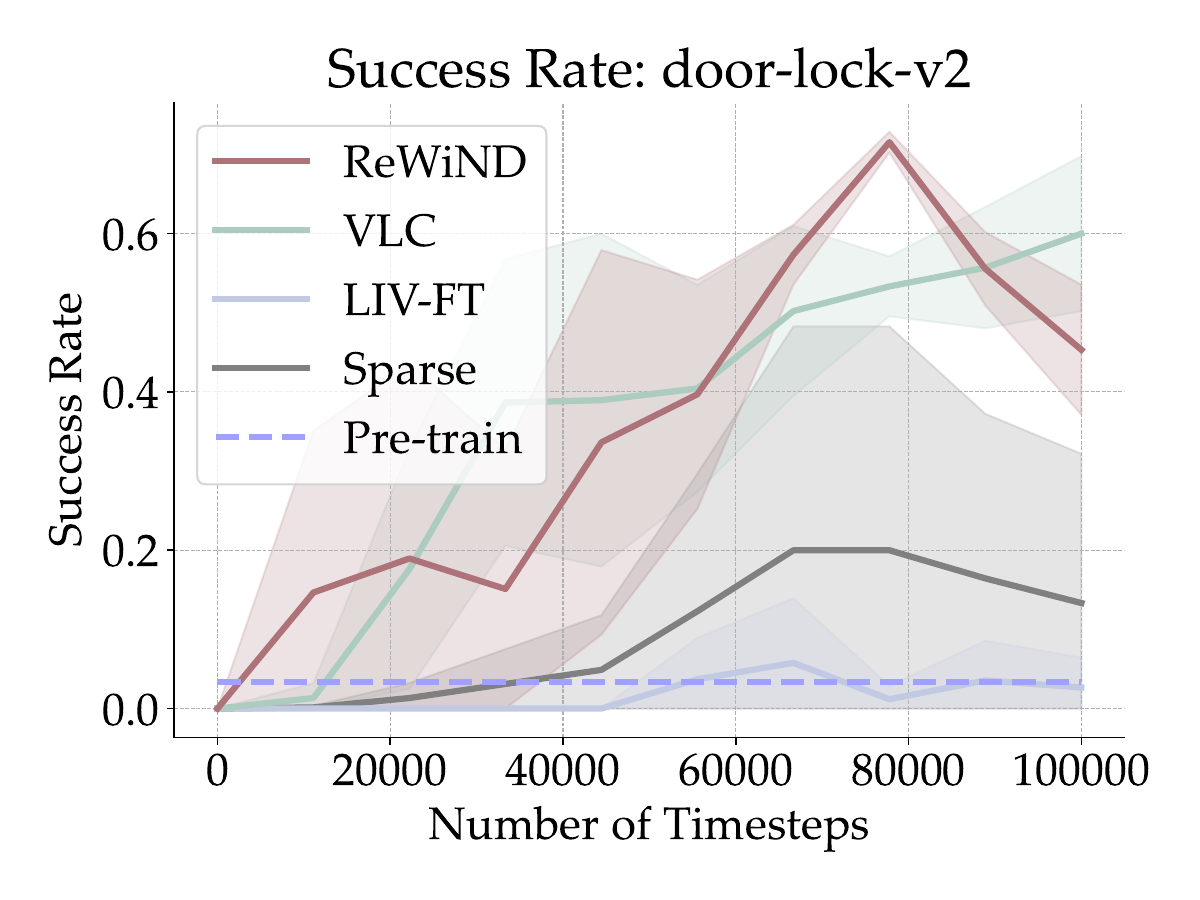}
    \caption{Door Lock}
  \end{subfigure}
  \hfill
  \begin{subfigure}[b]{0.32\textwidth}
    \includegraphics[width=\linewidth]{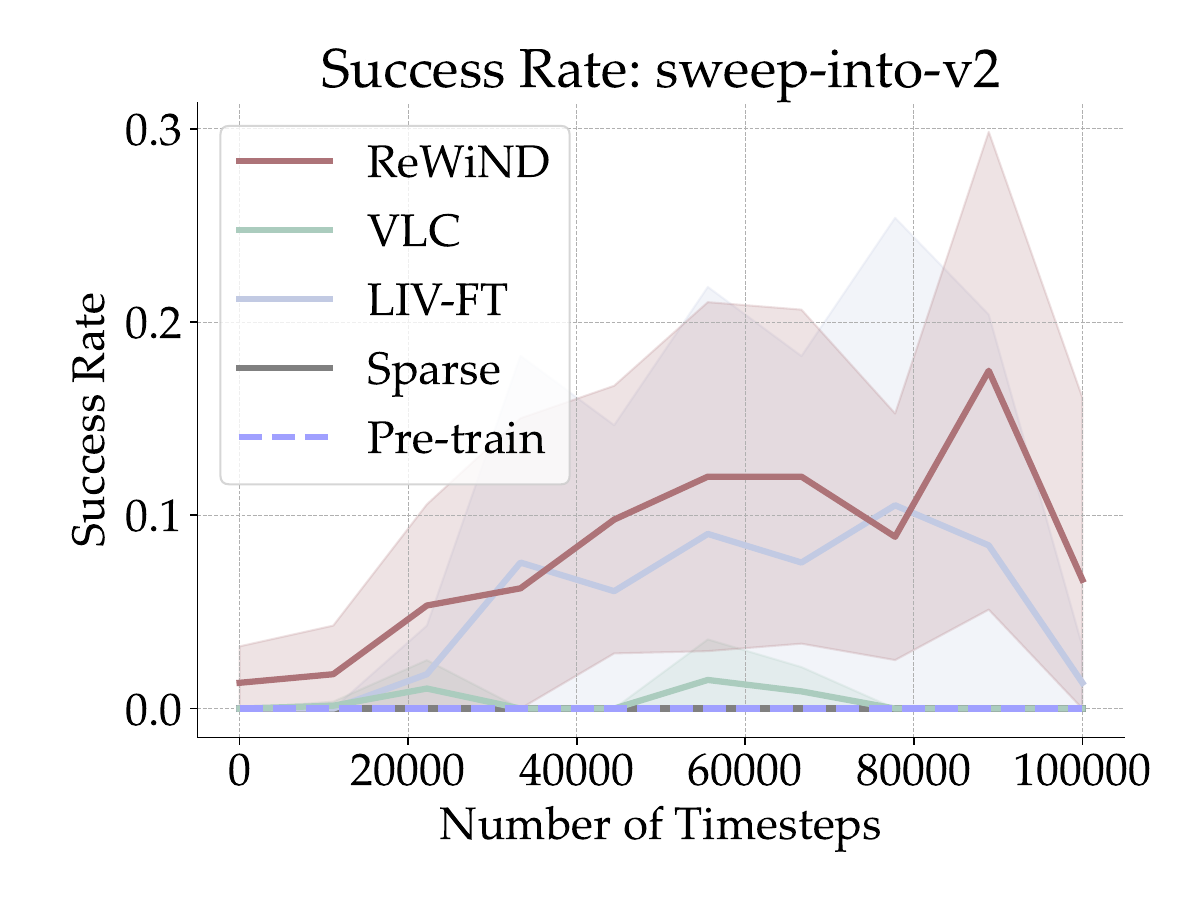}
    \caption{Sweep Into}
  \end{subfigure}
  \hfill
  \begin{subfigure}[b]{0.32\textwidth}
    \includegraphics[width=\linewidth]{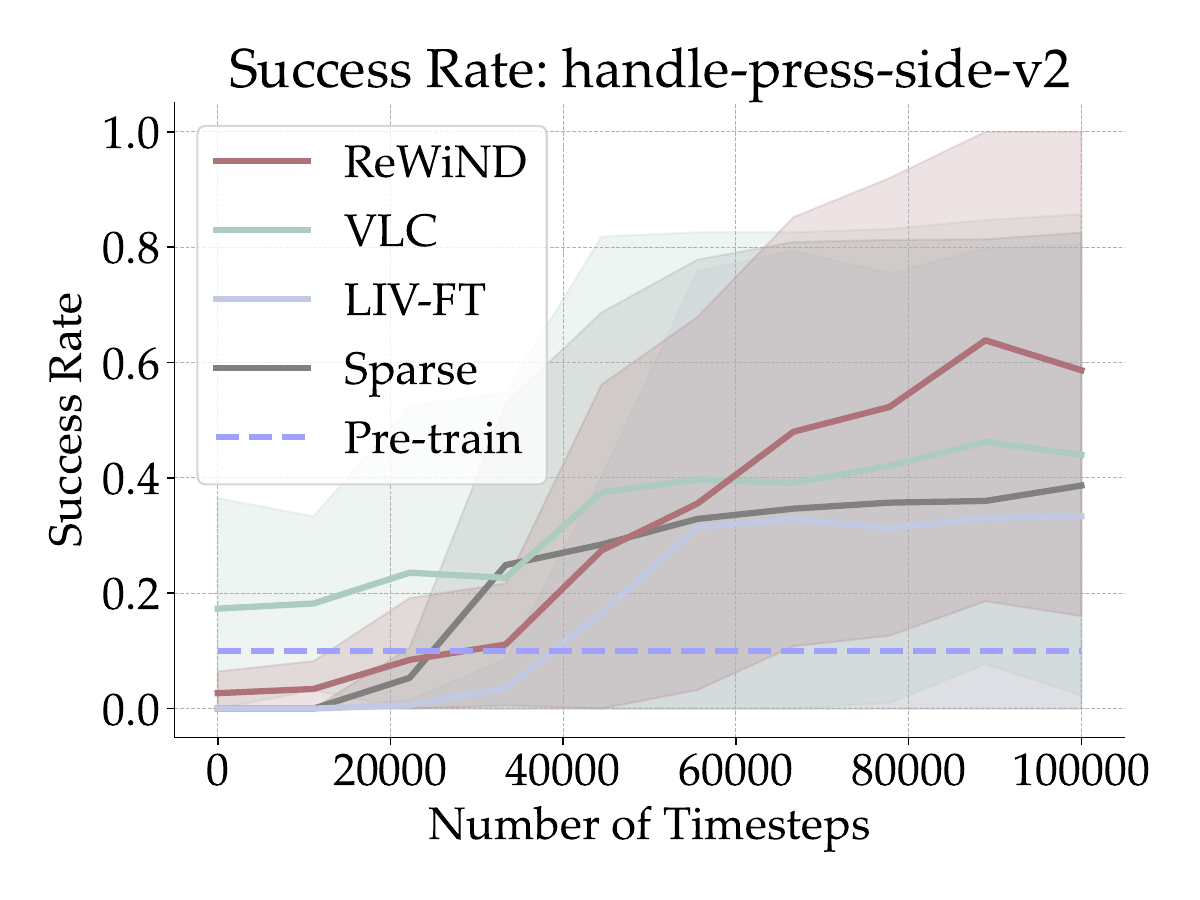}
    \caption{Handle Press (Side)}
  \end{subfigure}
  \hfill
  \begin{subfigure}[b]{0.32\textwidth}
    \includegraphics[width=\linewidth]{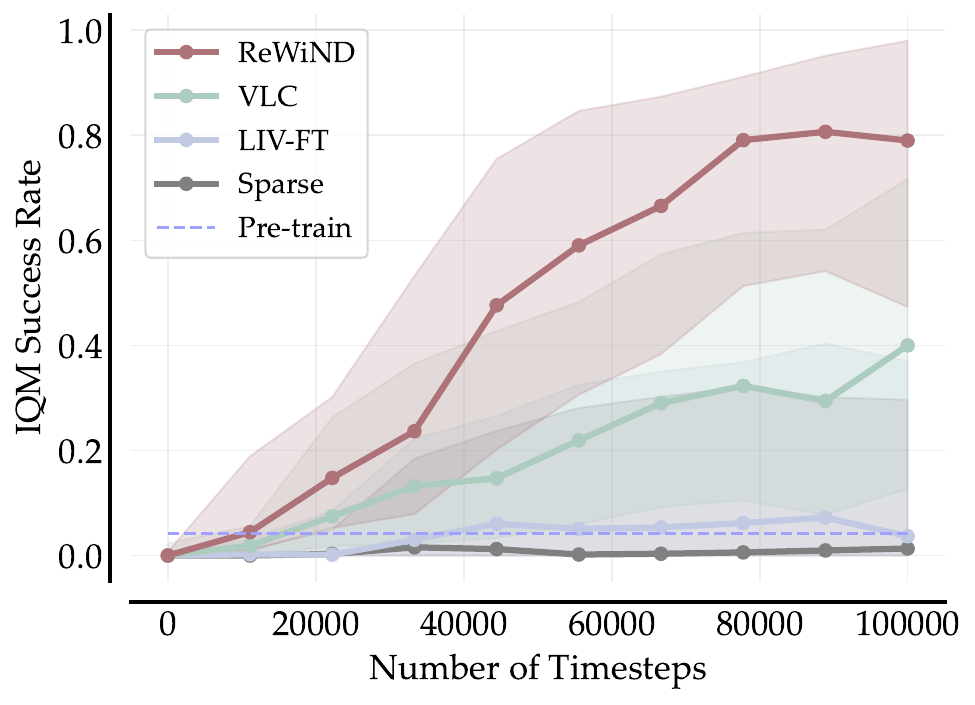}
    \caption{All Tasks IQM and 95\% CI}
    \label{fig:metaworld_curves:average}
  \end{subfigure}

  \caption{\textbf{MetaWorld success curves.} Task-level success rate learning curves plotting mean and shaded standard deviations. The bottom right figure plots the overall average across all tasks in terms of IQM and 95\% confidence intervals.}
  \label{fig:metaworld_curves}
\end{figure}

In this section, we analyze the sample efficiency of \method\ against baselines in MetaWorld. \Cref{fig:metaworld_curves} plots the learning curves for all downstream policy training tasks. Each panel corresponds to one specific task. And \Cref{fig:metaworld_curves:average} displays the average of all 8 downstream tasks we used for policy fine-tuning. We can see from the average IQM plot that \method\ achieves higher success rate than other baselines with the same number of timesteps and \method\ is generally more sample-efficient at any timestep.

\subsection{Real-World Reward Analysis}
\label{sec:appendix:additional_results:real_world_reward}

\begin{wraptable}{R}{0.6\linewidth}
\centering
\caption{\textbf{Evaluation Metrics on Real-world Unseen Tasks:} Comparsion between reward models in real-world unseen tasks with rank correlation $\rho$ and $r$.}
\resizebox{\linewidth}{!}{%
\begin{tabular}{@{}ccccccc@{}}
\toprule
Model & LIV  & LIV-FT & RoboCLIP & VLC  & GVL  & ReWiND \\ \midrule
  $\rho$ $\uparrow$    & 0.22 & -0.18  & 0.04     & 0.20 & 0.57 & \textbf{0.91}   \\
  $r$ $\uparrow$    & 0.23 & -0.13  & 0.04     & 0.19 & 0.52 & \textbf{0.91}   \\ \bottomrule
\end{tabular}
}
\label{tab:eval_koch_table}
\end{wraptable}

\begin{figure}[t]
    \centering
    \includegraphics[width=\linewidth]{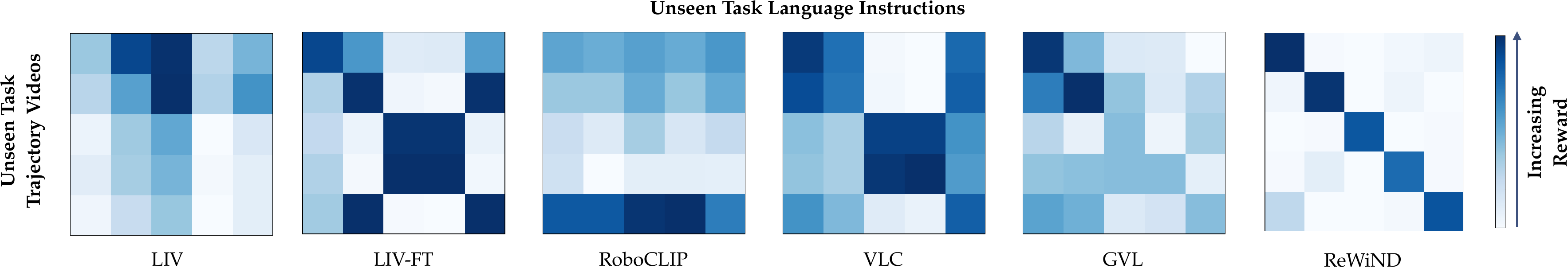}
    \vspace{-0.4cm}
        \caption{\textbf{Real-world Koch Reward Confusion Matrix on 5 Unseen Tasks.} 
        For each unseen task, we compute rewards for all combinations of demonstration videos and language descriptions. 
        \method\ produces the most \textcolor{blue}{diagonal-heavy} confusion matrix, indicating strong alignment between unseen demos and instructions.
        }
    \label{fig:eval_koch_confusion_matrices}
\end{figure}

\begin{figure}[t]
    \centering
    \includegraphics[width=\linewidth]{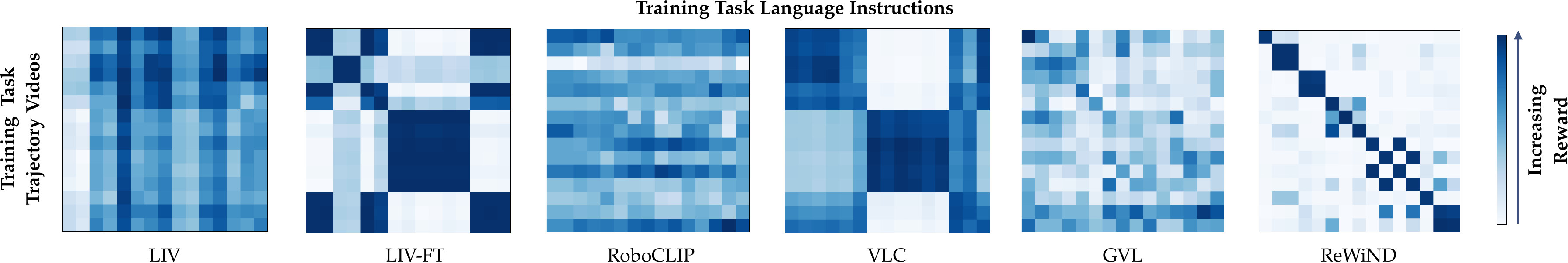}
    \vspace{-0.4cm}
        \caption{\textbf{Real-world Koch Reward Confusion Matrix on 15 Training Tasks.} 
        For each training task, we compute rewards for all combinations of demonstration videos and language descriptions. LIV-FT, VLC and \method\ are pretrained or fine-tuned with these training tasks while LIV , GVL and RoboCLIP are not
        }
    \label{fig:train_koch_confusion_matrices}
\end{figure}

We evaluated the performance of \method\ in MetaWorld in \Cref{sec:experiments:reward_analysis}. In this section, we analyze how \method\ works with real-world data.
\newtext{For the real-world setup, we use both views of each trajectory, treated as separate videos (but from the same demonstration) to train and evaluate all models.}
It can be seen from \Cref{tab:eval_koch_table} that \method\ has the highest Spearman’s rank correlation ($\rho$) and Pearson’s rank correlation ($r$) among all reward models. Also, in \Cref{fig:eval_koch_confusion_matrices} and \Cref{fig:train_koch_confusion_matrices}, \method\ has the best alignment between true-paired video and language instruction in both training tasks and unseen tasks, displaying strong generalization in new tasks. Note that LIV, GVL, and RoboCLIP are not trained on these training tasks as they are zero-shot models. 

\end{document}